%% file: incentive_MAB.tex
\documentclass{article}
\usepackage{microtype}
\usepackage{graphicx}
\pdfoutput=1
\usepackage{ifpdf}
\usepackage{booktabs} 

\usepackage{amsmath, bm}
\usepackage{amsthm}
\usepackage{amssymb}
\usepackage{mathtools}
\usepackage{thm-restate}
\usepackage{subcaption}
\usepackage{hyperref}
\usepackage{appendix}
\usepackage{wrapfig}
\usepackage{subfiles}
\usepackage[many]{tcolorbox}

\newtheorem{theorem}{Theorem}
\newtheorem{lemma}[theorem]{Lemma}
\newtheorem{definition}{Definition}

\newtheorem{fact}{Fact}

\newtheorem{remark}{Remark}

\definecolor{titlebg}{RGB}{72,72,72}
\definecolor{introbg}{RGB}{128,128,128}

\newtcolorbox{usecase}[1][]{
  breakable,
  enhanced,
  arc=0pt,
  outer arc=0pt,
  colframe=titlebg,
  colback=titlebg!05,
  overlay unbroken and first={
    \node[
      draw=titlebg,
      fill=titlebg,
      rotate=0,
      anchor=north west,
      text=white,
      font=\bfseries
    ]
    at (frame.north west)  
    {#1};
  }
}

\usepackage[accepted]{icml2021}

\icmltitlerunning{Incentivized Bandit Learning with Self-Reinforcing User Preferences}

\begin{document}

\twocolumn[
\icmltitle{Incentivized Bandit Learning with Self-Reinforcing User Preferences}



\icmlsetsymbol{equal}{*}

\begin{icmlauthorlist}
\icmlauthor{Tianchen Zhou}{one}
\icmlauthor{Jia Liu}{one}
\icmlauthor{Chaosheng Dong}{two}
\icmlauthor{Jingyuan Deng}{two}
\end{icmlauthorlist}

\icmlaffiliation{one}{Department of Electrical and Computer Engineering, The Ohio State University, Columbus, Ohio, USA}
\icmlaffiliation{two}{Amazon, Seattle, Washington, USA}

\icmlcorrespondingauthor{Tianchen Zhou}{zhou.2220@osu.edu}
\icmlcorrespondingauthor{Jia Liu}{liu@ece.osu.edu}

\icmlkeywords{Multi-armed bandit, online learning, incentivized bandit}

\vskip 0.3in
]



\printAffiliationsAndNotice{}  

\begin{abstract}
In this paper, we investigate a new multi-armed bandit (MAB) online learning model that considers real-world phenomena in many recommender systems: (i) the learning agent cannot pull the arms by itself and thus has to offer payments to users to incentivize arm-pulling indirectly; and (ii) if users with specific arm preferences are well rewarded, they induce a ``self-reinforcing'' effect in the sense that they will attract more users of similar arm preferences.
Besides addressing the tradeoff of exploration and exploitation, another key feature of this new MAB model is to balance reward and incentivizing payment.
The goal of the agent is to minimize the accumulative regret over a fixed time horizon $T$ with a low total payment.
Our contributions in this paper are two-fold: 
(i) We propose a new MAB model with random arm selection that considers the relationship of users' self-reinforcing preferences and incentives; 
and (ii) We leverage the properties of a multi-color P\'{o}lya urn with nonlinear feedback models to propose two MAB policies termed ``At-Least-$n$ Explore-Then-Commit'' and ``UCB-List.''
We prove that both policies achieve $O(\log T)$ expected regret with $O(\log T)$ expected payment over a time horizon $T$.
We conduct numerical simulations to demonstrate and verify the performances of these two policies and study their robustness under various settings.
\end{abstract}

\input{Sec1_Introduction/Introduction.tex}
\input{Sec2_RelatedWork/RelatedWork.tex}

\input{Sec3_Model/Model.tex}

\input{Sec4_Policies/Policies.tex}

\input{Sec5_Simulation/Simulation.tex}

\input{Sec6_Conclusion/Conclusion.tex}

\bibliography{ref.bib}
\bibliographystyle{icml2021}

\input{Sec_Proof/Proof.tex}

\end{document}

%% file: Sec1_Introduction/Introduction.tex

\section{Introduction}\label{introduction}

In many online e-Commerce platforms, there exists a self-reinforcing phenomenon, where the current user's behavior is influenced by the user behaviors in the past \citep{barabasi1999emergence, chakrabarti2005influence, ratkiewicz2010characterizing}, or an item is getting increasingly more popular as it accumulates more positive feedbacks.
For example, on a movie rental website, current customers tend to have more interest in Movie A that has $500$ positive reviews, compared with Movie B that only has $10$ positive reviews.
As an online learner, the e-Commerce service provider wants to identify the most profitable item in order to maximize the total profit in the long run.
In the literature, such an online profit maximization problem can often be modeled by the multi-armed bandit (MAB) framework \citep{berry1985bandit, bubeck2012regret}.
However, existing works on MAB that consider the self-reinforcing preferences remain quite limited (see, e.g., \citet{fiez2018multi, shah2018bandit}).
In fact, \citet{shah2018bandit} showed that the self-reinforcing preferences might render the classic UCB (upper confidence bound) policy \citep{auer2002finite} sub-optimal, and  new optimal arm selection algorithms are necessary.

On the other hand, in many online learning problems that utilize the MAB framework for  sequential decision making (e.g., recommender systems, healthcare, finance, dynamic pricing, see \citet{bouneffouf2019survey}), the learning agent (e.g., an online service provider) {\em cannot} select the arms directly.
Rather, arms are pulled by the users who are exhibiting self-reinforcing preferences.
The agent thus needs to {\em incentivize} users to select certain arms to maximize the total rewards, while avoiding incurring high incentive costs.
Hence, the bandit models in \citep{fiez2018multi, shah2018bandit} are no longer applicable, even though the self-reinforcing preferences behavior is considered.
Meanwhile, there exist several works \citep{frazier2014incentivizing, mansour2015bayesian, mansour2016bayesian, wang2018multi} that studied incentivized bandit under various settings and proposed efficient algorithms (more details in Section \ref{sec:related works}), but none of these works models users with self-reinforcing preferences.

The missing of joint modeling of incentives and self-reinforcing preferences in the existing MAB framework (two key features of many online e-Commerce systems) motivates us to fill this gap in this paper.
Specifically, in this work,
we first propose a more general MAB model with {\em stochastic arm selections following user preferences}, which is closely modeling random user behaviors in most online recommender systems.
This is in stark contrast to most existing works in the areas of incentivized bandits \citep{frazier2014incentivizing, wang2018multi}, where a (unrealistic) deterministic greedy user behavior is often assumed.
Under this model, a pair of fundamental trade-offs naturally emerge: (1) Sufficient exploration is required to identify an optimal arm, which may result in multiple pullings of sub-optimal arms, while adequate exploitation is needed to stick with the arm that did well in the past, which may or may not be the best choice in the long run;
(2) The agent needs to provide enough incentives to mitigate unfavorable initial bias and self-reinforcing user preferences, while in the meantime avoiding unnecessarily high incentives for users.
As in most online learning problems, we use regret as a benchmark to evaluate the performance of our MAB policy, which is defined as the performance gap between the proposed policy and an optimal policy in hindsight. 
The major challenges in this new MAB model thus lie in the following fundamental questions: 
\vspace{-.1in}
\begin{list}{\labelitemi}{\leftmargin=1.5em \itemindent=-0.0em \itemsep=-.1em}
\item[(a)] During incentivized pulling, how could the agent maintain a good balance between exploration and exploitation to minimize regret?

\item[(b)] How long should the agent incentivize until the right self-reinforcing user preference is established toward an optimal arm (so that no further incentive is needed)?

\item[(c)] Is the established self-reinforcing user preferences sufficiently strong and stable to sustain the sampling of an optimal arm over time without additional incentives? 
If yes, under what conditions could this happen?
\end{list}

\vspace{-.1in}
In this work, we answer the above questions by proposing two ``$\log(T)$-regret-with-$\log(T)$-payment'' policies for the incentivized MAB framework with self-reinforcing preferences.
Our contributions are summarized as follows:
\vspace{-.1in}
\begin{list}{\labelitemi}{\leftmargin=1em \itemindent=-0.0em \itemsep=-.1em}
\item We first show that no incentivized bandit policy can achieve a sub-linear regret with a sub-linear total payment if the feedback function that models the self-reinforcing preferences has a super-polynomial growth rate.
The proof is inspired by a multi-color P\'{o}lya urn model, and we also show how to guide the self-reinforcing preferences toward a desired direction.

\item To address the unique challenges in the new MAB model, we introduce (i) a {\em three-phase MAB policy architecture} and (ii) a key result that shows that an $O(\log T)$ incentivizing period is sufficient for establishing {\em dominance} for the multi-color P\'{o}lya urn model (see Section~\ref{sec:policies}).
All of these results are new in the bandit literature, which could be of independent interest for other incentivized MAB problems.

\item We propose two bandit policies, namely At-Least-$n$ Explore-Then-Commit and UCB-List, both of which are optimal in regret.
Specifically, for the two policies, we analyze the upper bounds of the expected regret and the expected total payment over a fixed time horizon $T$.
We show that both policies achieve $O(\log T)$ expected regrets, which meet the lower bound in \citet{lai1985asymptotically}.
Meanwhile, the expected total incentives for both policies are upper bounded by $O(\log T)$.

  \end{list}


%% file: Sec2_RelatedWork/RelatedWork.tex

\section{Related Work}\label{sec:related works}

The self-reinforcing phenomenon has received increasing interest in several different fields recently under different terminologies.
In the random network literature, previous works have studied the network evolution with ``preferential attachment'' \citep{barabasi1999emergence, chakrabarti2005influence, ratkiewicz2010characterizing}.
Also, a similar social behavior, referred to as \textit{herding}, is studied in the Bayesian learning model literature \citep{bikhchandani1992theory, smith2000pathological, acemoglu2011bayesian}.
For example, \citet{acemoglu2011bayesian} first studied the conditions under which there exists a convergence in probability to the desired action as the size of a social network increases.
More recently, \citet{shah2018bandit} incorporated positive externalities in user arrivals and proposed MAB algorithms to maximize the total reward.
Then, \citet{fiez2018multi} provided a more general model, where the learning agent has limited information.
We note that the agents in \citet{shah2018bandit, fiez2018multi} have full control in determining which arm for users to pull.
In contrast, the agent in our MAB model has {\em no control} over which arm to pull, and can only incentivize users to indirectly induce the preferences toward a desired arm.
Eventually, which arm to be pulled is entirely dependent on the current user's random preference.

On the other hand, incentivized MAB has attracted growing attention in recent years \citep{kremer2014implementing, frazier2014incentivizing, mansour2015bayesian, mansour2016bayesian, wang2018multi}.
To our knowledge, \citet{frazier2014incentivizing} first adopted incentive schemes into a Bayesian MAB setting.
In their model, the agent seeks to maximize time-discounted total reward by incentivizing arm selections.
\citet{kremer2014implementing} shares a similar motivation as \citet{frazier2014incentivizing}.
But in the model of \citet{kremer2014implementing}, the agent does not offer payments to the users.
Instead, he decides the information to be revealed to users as incentives.
Subsequently, \citet{mansour2015bayesian} studied the case where the rewards are not discounted over time.
More recently, \citet{wang2018multi} considered the non-Bayesian setting with non-discounted rewards.
\citet{agrawal2020incentivising} considered incentivizing exploration under contextual bandits.
These models differ from ours in both the incentive schemes and user behaviors.

Another line of research similar to incentivized bandit is bandit with budgets \citep{guha2007approximation, goel2009ratio, combes2015bandits, xia2015thompson}, where the agent takes actions with budget constraints.
\citet{guha2007approximation} developed approximation algorithms for a large class of budgeted learning problems.
Then, \citet{goel2009ratio} proposed index-based algorithms for this problem.
The key difference from our work is that in these models, the budget constraints are pre-determined, and the agents cannot take any further actions as soon as the budget constraints are violated.
In contrast, the total payment in our model is evaluated only after the time horizon is finished, which implies that bounding the total payment is part of our goals.

Although not cast in the MAB framework, the works on {\em urn models} \citep{khanin2001probabilistic, drinea2002balls, oliveira2009onset, zhu2009nonlinear} also share some relevant feedback settings to our model.
\citet{drinea2002balls} first proposed a class of processes called \textit{balls and bins models with feedback}, which is a preferential attachment model for large networks.
They then proved the convergence results of the model with various feedback functions.
Later, \citet{khanin2001probabilistic} improved the convergence result by showing monopoly (to be defined later) happens with probability one under a class of feedback functions included in \citet{drinea2002balls}.
Our proposed model is inspired by the ideas of feedback from \citet{oliveira2009onset}, in which the author discussed a natural evolution of the balls and bins process with nonlinear feedback.
However, our model is focused on MAB regret minimization, which is completely different from the goals considered in these works.

%% file: Sec3_Model/Model.tex

\section{System Model and Problem Statement}\label{sec:model}

\begin{figure}[t!]
	\centering
    \includegraphics[trim=0 8 0 0,width=\linewidth]{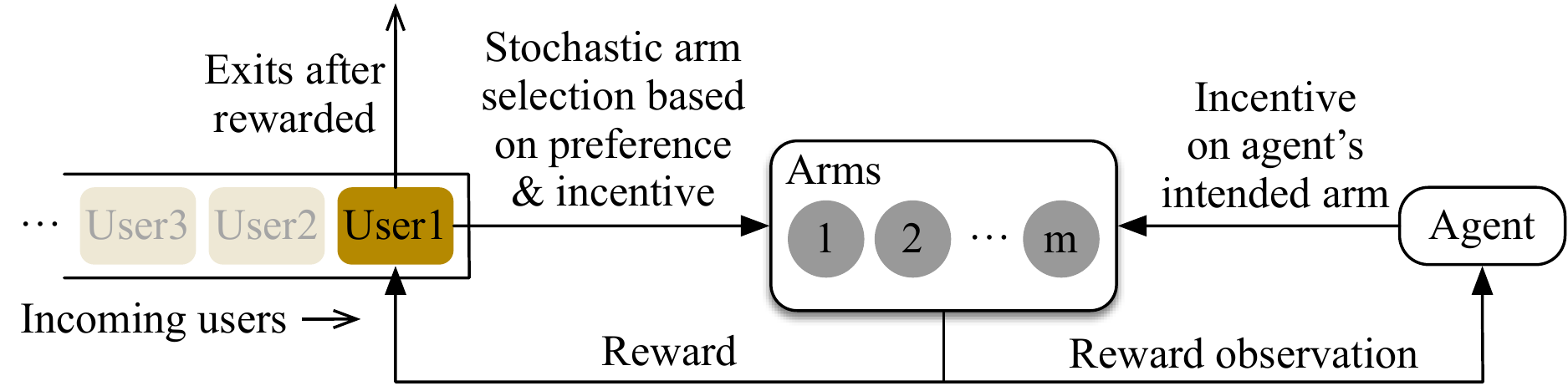}
    \caption{Incentivized MAB model with stochastic arm selection based on user preference rates and incentives.}
	\label{model}
\end{figure}

In this paper, we denote the set of arms offered by the agent as $A=\lbrace 1, \ldots, m\rbrace$.
Each arm $a$ follows a Bernoulli reward distribution $D_a$ with an unknown mean $\mu_a>0$.
The process runs for $T$ rounds.
As shown in Fig.~\ref{model}, in each time step $t \in \{1, \ldots, T\}$, a user arrives and chooses an arm $I(t)$ to pull, then receives a random reward $X(t)\sim D_{I(t)}$, which is observable to the agent.
We use $T_a(t)\triangleq \sum_{i=1}^t\mathrm{1}_{\lbrace I(i)=a\rbrace}$ to denote the number of times that an arm $a$ is pulled up to time $t$.
We denote the total reward generated by arm $a$ up to time $t$ as $S_a(t)\triangleq \sum_{i=1}^tX(i)\cdot\mathrm{1}_{\lbrace I(i)=a\rbrace}$.
We let $T_a(0)=0$ and $S_a(0)=0$, $\forall a \in A$.
We assume that there is a unique best arm $a^*\in A$, i.e., $a^* = \arg\max_a \mu_a$ and $\mu^*=\mu_{a^*}$.

\smallskip
{\bf 1) Preference and Bias Modeling:}
Unlike most of the incentivized MAB models where users are rational and independent, the user behavior is {\em stochastic} and {\em influenced by history} in our model.
Specifically, in each time step $t$, the user has a non-zero probability $\lambda_a(t) \in (0,1)$ to pull each arm $a\in A$, with $\sum_{a\in A}\lambda_a(t)=1, \forall t$.
In other words, the probability $\lambda_a(t)$ can be viewed as the {\em preference rate} of arm $a$ in time step $t$.
We adopt the widely used multinomial logit model in the literature to model $\lambda_a(t)$ as follows:
\begin{equation} \label{eqn_pref_model}
    \lambda_a(t) = \cfrac{F\big(S_a(t-1) + \theta_a\big)}{\sum_{i\in A}F\big(S_i(t-1) + \theta_i\big)},
\end{equation}
where $F(\cdot):\mathbb{R}\rightarrow (0, +\infty)$ is a feedback function that is increasing, and $\theta_a > 0$ denotes the fixed initial preference bias of arm $a$.
Intuitively, the increasing feedback function $F(\cdot)$ models the {\em self-reinforcing user preference effect} in the following sense: if an arm $a$ has been more profitable in the past, a user who prefers arm $a$ is more likely to arrive in the next round.
A simple example of the feedback function is $F(x) \!=\! x^\alpha$ for some constant $\alpha\!>\!1$. Here, $\alpha$ represents the strength of the self-reinforcing preference: a larger $\alpha$ implies a stronger self-reinforcing preference effect.

Several important remarks for the preference model in \eqref{eqn_pref_model} are in order.
The multinomial logit model is based on the behavioral theory of utility and has been widely applied in the marketing literature to model the brand choice behavior \citep{guadagni2008logit, gupta1988impact}.
The multinomial logit model is also used in the social network literature to model preferential attachment \citep{barabasi1999emergence}, where the probability that a link connects a new node $j$ with another existing node $i$ is linearly proportional to the degree of $i$.
Notably, this multinomial logit model has also been adopted in \citet{shah2018bandit} to model the same type of self-reinforcing phenomenon in their MAB model.

%
%

\smallskip
{\bf 2) Incentive Mechanism Modeling:}
Unlike in conventional MAB models,
the agent in our model can only offer some {\em incentive} on the arm that the agent wants to explore, so as to increase the users' preferences of pulling this particular arm for the agent (as shown in Fig.~\ref{model}).
The agent's goal is to maximize total reward in the long run.
%
In this paper, we model the influence of the incentives by adopting the so-called ``coupon effects on brand choice behaviors'' in the economics literature \citep{papatla1996measuring, bawa1987effects}.
In this model, the relationship between coupons and choices is nonlinear, and the redemption rate increases with respect to the coupon value but exhibits a diminishing return effect \citep{bawa1987effects}.
Specifically, in time step $t$, if the agent wants to explore arm $a$, the agent will offer a fixed payment $b$\footnote{In this paper, we consider fixed payment with the goal of gaining a first fundamental understanding of the regret of the proposed new MAB model.
The problem of optimizing the total cost of a time-varying payment strategy is an important related problem, which will left for our future studies.}
to the current user to increase the user's preference on pulling arm $a$.
Under the coupon effect model, the posterior preference rates of the arms with incentive $b$ are updated as follows:
\begin{equation} \label{eqn_pref_update}
\!\!\!    \hat\lambda_i(t) \!=\! \begin{cases}
    \cfrac{\bar{G}(b,t) + F\big(S_i(t-1) + \theta_i\big)}{\bar{G}(b,t) + \sum_{j\in A}F\big(S_j(t-1) + \theta_j\big)}, & \!\!\! i=a,\\[1em]
    \cfrac{F\big(S_i(t-1) + \theta_i\big)}{\bar{G}(b,t) + \sum_{j\in A}F\big(S_j(t-1) + \theta_j \big)}, & \!\!\! i\neq a,
\end{cases} \!\!\!
\end{equation}
where $\bar{G}:\mathbb{R}^2 \rightarrow \mathbb{R}^{+}$ is an increasing function of $b$ with $\bar{G}(0,\cdot)=0$, which can be interpreted as the impact of payment $b$ on users at time $t$.
Intuitively, $\bar{G}(b,t)$ represents the ``impact'' of offering incentive $b$ on users at time $t$.
Also, $\bar{G}(b,t)$ has the property that it is increasing over time.
The interpretation is that, as arms gain higher accumulative total reward $\sum_{i\in A}F\big(S_i(t-1) + \theta_i\big)$ as $t$ increases (e.g., items gaining more positive reviews), offering the same amount of incentive $b$ on any of them becomes more attractive.

Clearly, the posterior preference update in (\ref{eqn_pref_update}) still follows the multinomial logit model.
Also, we can see from (\ref{eqn_pref_update}) that, as parameter $b$ increases asymptotically ($b \uparrow \infty$), we have $\hat{\lambda}_{a}(t) \uparrow 1$ and $\hat{\lambda}_{i}(t) \downarrow 0$, $\forall i\ne a$, i.e., arm $a$ is preferred with probability one.
For simplicity in our subsequent analysis, in the rest of the paper, we rewrite $\hat\lambda_i(t)$ in the following equivalent form: we divide both the denominator and numerator by $\sum_{i\in A}F\big(S_i(t-1) + \theta_i\big)$ and let $G(b,t) \triangleq \bar{G}(b,t)/\sum_{i\in A}F\big(S_i(t-1) + \theta_i\big)$. 
Then, it can be verified that Eq.~\eqref{eqn_pref_update} can be equivalently rewritten as:
\begin{equation*}
    \hat\lambda_i(t) = \begin{cases}
    \cfrac{\lambda_i(t) + G(b,t)}{1 + G(b,t)}, &i=a,\\[1em]
    \cfrac{\lambda_i(t)}{1 + G(b,t)}, & i\neq a.
\end{cases}
\end{equation*}
Clearly, $G(b,t)$ remains an increasing function of $b$.
Also, we define the accumulative payment up to time step $t$ as $B_t\coloneqq \sum_{i=1}^t b_t$, where $b_{t} \in \{ 0, b \}$, $\forall t$, denotes the agent's binary decision whether to offer incentive $b$ at time step $t$.

{\bf 3) Regret Modeling:}
Let $\Gamma_T = \sum_{t=1}^T X(t)$ denote the accumulative reward up to time $T$.
In this paper, we aim to maximize $\mathbb{E}[\Gamma_T]$ by designing an incentivized policy $\pi$ with low accumulative payment in terms of growth rate with respect to $T$.
A policy $\pi$ is an algorithm that produces a sequence of arms that are recommended at time step $t = 1, \ldots, T$.
Similar to conventional MAB problems, we measure our accumulative reward performance against an oracle policy, where in hindsight the agent knows the best arm $a^*$ with the largest mean and can always offer an {\em infinite} amount of payments to users, so that the updated preference rate of arm $a^*$ is always infinitely close to one.
We denote the expected accumulative reward generated under the oracle policy up to time $T$ as $\mathbb{E}[\Gamma^*_T] = \mu_{a^*}T$.\footnote{It is insightful to compare our oracle policy with \citet{shah2018bandit}. The oracle policy in \citet{shah2018bandit} does not achieve $\mu_{a^*}T$ expected accumulative reward up to time $T$ due to the following key modeling difference: In \citet{shah2018bandit}, it is assumed that the agent can only feed a {\em single arm} at a time to the current user.
Hence, the oracle policy keeps {\em only} feeding the best arm to all arriving users.
However, in the early time steps, a fraction of the users may not prefer the best arm due to initial biases.
Hence, the agent has to spend time mitigating these initial biases, resulting in an expected accumulative reward smaller than $\mu_{a^*}T$. 

In contrast, we assume that the agent can feed {\em all arms} to each user (closely models real-world recommender systems), and the oracle policy offers an infinite amount of payment as incentives.
As a result, users will always pull the best arm with probability one in each time step, which implies $\mu_{a^*}T$ expected accumulative reward up to time $T$.
}
The expected (pseudo) regret is defined as: $\mathbb{E}[R_T] = \mu_{a^*}T - \mathbb{E}[\Gamma_T]$.
Our goal is to minimize $\mathbb{E}[R_T]$, with low expected accumulative payment $\mathbb{E}[B_T]$ with respect to the time horizon $T$.

%% file: Sec4_Policies/Policies.tex

\section{Policy Designs and Performance Analysis}\label{sec:policies}

In this section, we present two policies that achieve $O(\log T)$ expected regret with $O(\log T)$ accumulative payment with respect to time horizon $T$.

\subsection{The Basic Idea}
The main idea of our two proposed policies is based on a unique {\em three-phase MAB policy architecture}: 1) We first perform exploration among all arms by incentivizing pulling until we know the best-empirical arm is optimal, i.e., $\hat a^* = a^*$ with high confidence; 
2) We keep incentivizing the pulling of the best-empirical arm $\hat a^*$ until it dominates and attracts users who favor this arm;
and 3) We stop incentivizing and rely on the self-reinforcing user preference to continue pulling the optimal arm.
The success of our incentivized policy designs relies on guaranteeing the {\em dominance} of arm $\hat a^*$, which is defined as follows:

\begin{definition}[Dominance] \label{defn_dominance}
An arm is said to be dominant if it produces at least half of the total reward.
\end{definition}

Our MAB policy designs are based on a {\em key fact} that, if the feedback function $F(x)$'s growth rate is superlinear polynomial, then as soon as dominance is established, we can stop incentivizing and rely on the users' self-reinforcing preferences to converge to one arm within a finite number of rounds, i.e., an arm $a\in A$ is the only arm to be sampled eventually.
We call this event as the \textit{monopoly by arm $a$} ($mono_a$ for short).
We point out that a {\bf key contribution} in this paper is the insight that dominance happens {\em much sooner than} establishing monopoly (to be shown later that this only takes $O(\log(T))$ rounds). 
This fact further implies the existence of an incentivized policy with {\em sub-linear} total payment.
We formally state this fact as follows:

\begin{restatable}{lemma}{RestateLEMMA}{\em (Monopoly)}
\label{prop}
There exists an incentivized policy that induces users' preferences to converge in probability to an arm over time with sub-linear payment, if and only if $F(x)$ satisfies $\sum_{i=1}^{+\infty}\big(1/F(i)\big)<+\infty$.
\end{restatable}

\vspace{-.1in}
\begin{proof}[Proof Sketch of Lemma~\ref{prop}]
Our main technique for proving Lemma~\ref{prop} is an improved exponential embedding method.
This method simulates the reward generating sequence by random exponentials.
In what follows, we outline the key steps of the proof and relegate the details to the supplementary material.

{\em Step 1) Construction of an Equivalent Reward Generating Sequence:} Define a sequence $\lbrace\chi_j\rbrace_{j=1}^\infty$ denoting the reward generating order, where each element denotes the arm index.
Note that an arm index appears in $\lbrace\chi_j\rbrace$ only if it is pulled and generates a unit reward.
We want to construct a sequence $\lbrace \zeta_j\rbrace$ that has the same conditional distribution as $\lbrace\chi_j\rbrace$ given history $\mathcal{F}_{j-1}$.
Then, the constructed sequence $\lbrace \zeta_j\rbrace$ will be leveraged to prove the lemma.

For arm $i$, consider a collection of independent exponential random variables $\lbrace r_i(n)\rbrace$ such that $\mathbb{E}[r_i(n)]=1/[\mu_iF(n+\theta_i)]$.
We construct an infinite set $B_i=\lbrace\sum_{k=0}^nr_i(k)\rbrace_{n=0}^\infty$, where each element $\sum_{k=0}^nr_i(k)$ models the time needed for arm $i$ to obtain accumulative reward $n$.
Then we mix and sort $B_i$ in an increasing order for all $i\in A$ to form a new sequence $H$.
Our objective sequence $\lbrace \zeta_j\rbrace$ is the arm index sequence out of $H$.
Then, we can prove by induction that given the previous reward history $\mathcal{F}_{j-1}$, the constructed sequence $\lbrace \zeta_j\rbrace$ has the same conditional distribution as $\lbrace\chi_j\rbrace$.

{\em Step 2) Establishing Attraction Time:}
The proof of Lemma~\ref{prop} is done once we show that if and only if any feedback function $F(x)>0$ satisfies $\sum_i \big(1/F(i)\big)<+\infty$, then $\mathbb{P}(\exists a\in A,\, mono_a)=1$.
We define the \textit{attraction time} $N$ as the time step when the monopoly happens.
With the constructed sequence $\lbrace \zeta_j\rbrace$, we establish the necessity by showing that if $\sum_i \big(1/F(i)\big)<+\infty$ then $\mathbb{P}(N<\infty)=1$, and the sufficiency by showing that if $\sum_i \big(1/F(i)\big)=+\infty$ then $\mathbb{P}(N=\infty)>0$.
This completes the proof.
\end{proof}

\begin{remark}{\em
	The exponential embedding technique has been applied in the literature (see, e.g., \citet{zhu2009nonlinear, oliveira2009onset, davis1990reinforced, athreya1968embedding}).
	This technique embeds a discrete-time process into a continuous-time process built with exponential random variables.
	We adapt it to our model by using exponential random variables with specific distributions.
	The most significant feature of our exponential embedding technique is that the random times of different arms generating unit rewards are independent and can be mathematically expressed as exponential distributions, which facilitates our subsequent analysis.}
\end{remark}

\begin{remark}{\em
	A simple example that satisfies the condition in Lemma~\ref{prop} is $F(x)= Cx^\alpha$ for some constants $C>0$ and $\alpha>1$ (i.e., superlinear polynomial). 
	In this case, there exists an incentivized policy that induces all preferences to converge over time with sub-linear total payment, since $\sum_{i=1}^{+\infty}(1/i^\alpha)<+\infty$ with $\alpha >1$.
	Previous works \citep{drinea2002balls,khanin2001probabilistic} considering the balls and bins model also studied this feedback function with $\alpha\leq1$.
	For $\alpha<1$, the asymptotic preference rates of arms are all deterministic, positive, and dependent on the means and biases of arms.
	For $\alpha=1$, the system is akin to a standard P\'{o}lya urn model, and will converge to a state where all arms have random positive preference rates depending on the means and initial biases of the arms.
	For $\alpha>1$, the system converges almost surely to a state where only one arm has a positive probability to generate rewards, depending on the means and initial biases of arms.
	Thus, systems under these three $\alpha$-values exhibit completely different behaviors.}
\end{remark}

\begin{remark}{\em
In our later theoretical and numerical studies in this paper, we will focus on the class of polynomial functions $F(x) = \Theta(x^{\alpha})$ with $\alpha > 1$ as the feedback function.
We note that the use of $F(x) = \Theta(x^{\alpha})$ does not lose much generality since all analytic functions in a bounded range can be approximated arbitrarily well by their Taylor polynomial expansions.
Also, since $F(x)$ that satisfies the condition $\sum_{i=1}^{+\infty}\big(1/F(i)\big)<+\infty$ in Lemma~\ref{prop} is lower bounded by $\Omega(x^\alpha)$ with $\alpha>1$ (by considering $\sum_{i=1}^{+\infty}\big(1/F(i)\big)$ as $p$-series), $F(x)=\Theta(x^\alpha)$ with $\alpha>1$ is general enough to cover a large class of functions.
}
\end{remark}

\subsection{The At-Least-$n$ Explore-Then-Commit Policy} \label{subsec:etc}
Our first policy is the At-Least-$n$ Explore-Then-Commit (AL$n$ETC), which consists of three phases: the exploration phase, the exploitation phase, and the self-sustaining phase.
The agent incentivizes in the first two phases.
During the exploration phase, AL$n$ETC explores all arms  until each arm generates sufficient accumulative reward.
Then, the policy incentivizes the arm with the best empirical mean until it {\em dominates} (as defined in Definition~\ref{defn_dominance}).
Toward this end, we define the sample mean of arm $a\in A$ at time step $t$ as $\hat\mu_a(t) = S_a(t-1)/T_a(t-1)$. 
Then, we formally state the AL$n$ETC policy as follows:

\begin{usecase}[Policy~1: At-Least-$n$ Explore-Then-Commit]
\vspace{0.15in}
Given time horizon $T$, payment $b$ and $n = q\ln T$, where $q>0$ is some tuning parameter:

\smallskip
{\normalfont \textbf{1) Exploration Phase:}} 
Incentivize pulling arm $a\in \arg\min_{i\in A}S_i(t)$ with payment $b$ until time $\tau_n = \min\lbrace t: S_a(t)\geq n,\, \forall a\rbrace\land T$, when any arm has accumulative reward of at least $n$.

\smallskip
{\normalfont \textbf{2) Exploitation Phase:}}
Incentivize pulling the best-empirical arm $\hat a^*\in \arg\max_{a\in A}\hat\mu_a(\tau_n)$ with payment $b$ until it dominates, i.e., $S_{\hat a^*}(t)\geq\sum_{a\neq \hat a^*}S_a(t)$. Mark current time as $\tau_s$.

\smallskip
{\normalfont \textbf{3) Self-Sustaining Phase:}}
Users pull arms based on their own preferences until time $T$.

\end{usecase}

%
%
%


For the AL$n$ETC policy, we next show that if the incentive effect is sufficiently strong, then the dominance time $\tau_s$ happens within $O(\log T)$ rounds, which is much sooner than the attraction time (i.e., time for establishing monopoly).
We formally state this result as follows:

\begin{restatable}{lemma}{RestateLEMMAD}{\em (Dominance)}
\label{dominance}
In AL$n$ETC, if the incentive sensitivity function $G(\cdot)$ and the payment $b$ satisfy $G(b,t)>1$ for all $t$ in the exploration and exploitation phases, then the expected dominant time $\tau_s$ is $O(\log T)$.
\end{restatable}

\begin{remark} {\em
In Lemma~\ref{dominance}, the condition ``$G(b,t)>1$'' has an interesting interpretation in practice.
Recall that $G(b,t)$ is defined as $G(b,t) \triangleq \bar{G}(b,t)/\sum_{i\in A}F\big(S_i(t-1) + \theta_i\big)$ (cf. Section~\ref{sec:model}).
Thus, $G(b,t)\!>\! 1$ means that the ``incentive impact'' $\bar{G}(b,t)$ should be larger (could be ever so slightly) than the ``impact of arms' accumulative reward'' $\sum_{i\in A} F(S_i(t\!-\!1)\!+\!\theta_i)$ so that incentive control is possible.
}
\end{remark}

Based on the above result, we will show next that once the best-empirical arm {\em dominates}, then it implies sub-linear regret and accumulative incentive payment.
Intuitively, this is because we will show that, within a finite number of steps after dominance time $\tau_s$, monopoly happens with probability one, and arm $\hat a^*$ has a high probability to emerge victorious in the monopoly (to be shown in the proof of Theorem \ref{thm-etc}). 
If the time horizon $T$ is sufficiently large to cover the attraction time (i.e., the time when monopoly happens), then arm $\hat a^*$ will be sampled repeatedly after the attraction time, while the expected pulling times from sub-optimal empirical arms after the dominance is $o(\log T)$ (which contributes to the regret).
Thus, the policy achieves a sub-linear expected regret.
For each arm $a$, we set $\Delta_a=\mu^*-\mu_a$, and let $\Delta_{min}=\min_{a\neq a^*}\Delta_a$, $\Delta_{max}=\max_{a\neq a^*}\Delta_a$.
We formally state this result as follows:

\begin{restatable}{theorem}{RestateETC}{\em (At-Least-$n$ Explore-Then-Commit)}
\label{thm-etc}
Given a fixed time horizon $T$, if (i) $G(b,t)>1$, (ii) $q\geq (2\max_{a\neq a^*}\mu_a)/\Delta_{min}^2$, (iii) $F(x)=\Theta(x^\alpha)$ with $\alpha>1$, then the expected regret of AL$n$ETC  is upper bounded by:
\begin{equation*}
	\mathbb{E}[R_T] \leq \sum\limits_{a\in A}\cfrac{2(G(b,t)-L_{a^*})\Delta_{max}}{\big(G(b,t)-1\big)\mu_a}\cdot q\ln T + o(\log T),
\end{equation*}
where $L_a=F(q\ln T+\theta_a)/\sum_{i\in A}F(\mu^*T+\theta_i)$.
The expected total payment is upper bounded by:
\begin{equation*}
	\mathbb{E}[B_T]\leq\sum\limits_{a\neq a^*}\cfrac{2b(G(b,t)+1)}{\mu_a(G(b,t)-1)}\cdot q\ln T.
\end{equation*}
\end{restatable}

\begin{remark}{\em
	For a given incentive $b$, as $G(b,t)$ increases asymptotically (large incentive impact), regret and total payment decrease to some limiting amounts.
	This makes intuitive sense since if the incentive has a larger impact on users, it will reduce the pullings of random unfavorable arms and shorten the exploration and exploitation phases. 
	On the other hand, as $G(b,t)$ decreases towards one from above, users are less affected by incentives, thus in many instances the exploration phase never stops.
	This could lead to linear expected regret and linear expected total payment.
	Meanwhile, as $q$ decreases, both regret and total payment are smaller.
	But if $q<(2\max_{a\neq a^*}\mu_a)/\Delta_{min}^2$, the exploration will be insufficient to guarantee the event $\lbrace\hat a^*=a^*\rbrace$.
	This leads to a linear regret.
	Also, a large $\Delta_{max}$ implies larger a loss of pullings of suboptimal arms to reach $n$ accumulative reward during exploration phase, leading to a larger regret.
	}
\end{remark}

\begin{proof}[Proof Sketch of Theorem~\ref{thm-etc}]
Due to space limitation, we provide a proof sketch here and relegate the details to the supplementary material.
By the law of total expectation, the expected regret up to time $T$ can be decomposed as:
\begin{equation*}
	\mathbb{E}[R_T]\leq\underbrace{\mathbb{E}[R_T \mid\hat a^* = a^*]}_\text{(a)} + T\cdot\underbrace{\mathbb{P}(\hat a^*\neq a^*)}_\text{(b)}.
\end{equation*}
To bound $\mathbb{E}[R_T]$, we want to upper bound both $\mathbb{E}[R_T \mid\hat a^* = a^*]$ and $\mathbb{P}(\hat a^*\neq a^*)$.
First, in (b), the probability $\mathbb{P}(\hat a^*=a^*)\leq\mathbb{P}\big(\hat\mu_a(\tau_n)\geq \hat\mu_{a^*}(\tau_n)\big)$ is bounded by $O(T^{-1})$ by leveraging the Chernoff-Hoeffding bound.
Also, noting that
\begin{equation*}
	(a)=\mu^* T - \big( \mathbb{E}[\Gamma_{\tau_s}\mid\hat a^* = a^*] + \mathbb{E}[\Gamma_T-\Gamma_{\tau_s}\mid\hat a^* = a^*]\big),
\end{equation*}
where $\Gamma_t$ is the accumulative reward up to time $t$, we first need to upper bound $\mathbb{E}[\tau_n]$ and $\mathbb{E}[\tau_s]$.
Consider $\mathbb{E}[\tau_n]$, we show that the number of pulling of arm $a$ to get a unit reward is a geometric random variable with parameter larger than $\mu_aG(b,t)/\big(G(b,t)+1\big)$.
Then, for each arm $a\in A$ to obtain at least $n$ accumulative reward, the expected time needed is upper bounded by
\vspace{-3pt}
\begin{equation*}
	\mathbb{E}[\tau_n] \leq \cfrac{G(b,t)+1}{G(b,t)}\cdot \sum_{i\in A}\cfrac{q\ln T}{\mu_i}.
\end{equation*}
For $\mathbb{E}[\tau_s]$, since $\tau_s$ is the earliest time for the system to reach dominance, $\tau_s$ satisfies the condition $\mu_{\hat a^*}\mathbb{E}[T_{\hat a^*}(t)] \geq \sum_{a\neq \hat a^*}\mu_a\mathbb{E}[T_a(t)]$.
With the bound of $\mathbb{E}[\tau_n]$, after relaxing the inequality and some rearrangement, we obtain the upper bound as follows:
\vspace{-3pt}
\begin{equation*}
	\mathbb{E}[\tau_s] \leq \cfrac{G(b,t)+1}{G(b,t)-1}\cdot\sum_{a\neq a^*}\cfrac{2q\ln T}{\mu_a}.
\end{equation*}
According to the policy, the expected accumulative payment $\mathbb{E}[B_T]$ can be bounded by $b\mathbb{E}[\tau_s]$ and part of the expected regret $\mathbb{E}[\Gamma_{\tau_s}\mid\hat a^* = a^*]$.

The next challenge is to show whether the dominant arm has a large enough probability to ``win'' in monopoly during the self-sustaining phase.
We use $D(u_0, n_0)$ to denote the ``bad event'' that the fraction of accumulative reward from weak arms increases over time. 
Formally, suppose that at time step $\tau_s$, there are $u_0n_0$ accumulative reward generated by weak arms, where $n_0$ is the total reward and $u_0<1/2$ is the fraction.
Then, $D(u_0, n_0)$ happens if $\exists t' \in (\tau_s,\, T]$, $un$ accumulative reward is generated from weak arms with fraction $u>u_0$.
The probability of event $D(u_0, n_0)$ can be bounded as $\mathbb{P}\big(\exists n > n_0,\, D(u_0, n_0)\big) \leq e^{-(u_0n_0)^\gamma}=e^{-O(\log T)^\gamma}$ with constant $\gamma\in(0,1/4)$ using the improved exponential embedding method and a Chernoff-like bound developed in the supplementary material. 
The upper bound of event $D(u_0, n_0)$ decreases as $u_0n_0$ increases monotonically over time.
Thus, the arms that stay on the weak side for a long time have little chance to win back.

Lastly, we bound the term $\mathbb{E}[R_T - R_{\tau_s}\hspace{-2pt}\mid\hspace{-2pt}\hat a^* = a^*]$ in $(a)$, which contributes to the $o(\log T)$ regret term in Theorem~\ref{thm-etc}.
After time $\tau_s$, a unit reward is generated by sub-optimal arms with probability upper bounded by $e^{-(u_0n_0)^\gamma}$, and then the next unit reward is also generated by sub-optimal arms with probability upper bounded by $e^{-(u_0n_0+1)^\gamma}$.
Thus, 
\begin{equation*}
	\mathbb{E}[R_T - R_{\tau_s}\mid \hat a^* = a^*]\leq e^{-(u_0n_0)^\gamma} + e^{-(u_0n_0+1)^\gamma} + \cdots,
\end{equation*}
with the summation on the right hand side bounded by $O\big((\log T)^{1-\gamma}e^{-(\log T)^\gamma}\big)$ and $\gamma\in(0,1/4)$.
\end{proof}

\subsection{The UCB-List Policy} \label{subsec:ucb-list}
In this section, we propose a UCB-List policy to further improve the performance of the AL$n$ETC policy.
UCB-List is similar to AL$n$ETC and also consists of three phases.
During the exploration phase, the agent initially puts all arms in one set, and then incentivizes the least pulled arm in the set.
Meanwhile, it removes arms that are estimated to be sub-optimal, until only one arm is left in the set, which is viewed as the best-empirical arm.
Note that in this phase, users can still pull any arm regardless of the set.
Then, the agent incentivizes users to sample the best-empirical arm until it dominates.
The UCB-list policy is stated as follows:

\begin{usecase}[Policy~2: The UCB-List Policy]
\vspace{0.15in}
Given time horizon $T$ and payment $b$, define the confidence interval of arm $a$ at time step $t$ as $c_a(t) = \sqrt{\ln T/2T_a(t)}$:

\smallskip
{\normalfont \textbf{Initialization:}} Incentivize pulling arms satisfying $T_a(t)=0$ with payment $b$ until $\min_{a\in A}T_a(t)=1$.
Let set $U = A$.

\smallskip
{\normalfont \textbf{1) Exploration Phase:}} While $|U|\!>\!1$, keep removing any arm $a$ satisfying $\hat\mu_a(t)+c_a(t) \leq \max_{i\neq a, i\in U}\big(\hat\mu_i(t)-c_i(t)\big)$ from $U$ if there is any.
Then, incentivize pulling arm $a\in \arg\min_{i\in U}T_i(t)$ with payment $b$.
If $|U|=1$, let arm $\hat a^* = \lbrace a: a\in U\rbrace$ and mark current time as $\tau_1$.

\smallskip
{\normalfont \textbf{2) Exploitation Phase:}} Incentivize pulling arm $\hat a^*$ with payment $b$ until it dominates: $S_{\hat a^*}(t)\geq\sum_{a\neq \hat a^*}S_a(t)$. Mark current time as $\tau_s$.

\smallskip
{\normalfont \textbf{3) Self-Sustaining Phase:}}
Users pull arms based on their own preferences until time $T$.

\end{usecase}

%
%
%
%

	Compared to AL$n$ETC that requires a tuning parameter $q$, UCB-List does not need any tuning parameter and dynamically eliminates suboptimal arms, while still balancing the exploration-exploitation trade-off to achieve $O(\log(T))$ regret and $O(\log(T))$ payment.
	We state this result as follows: 

\begin{restatable}{theorem}{RestateUCB}{\em (UCB-List)}
\label{thm-ucb-list}
Given a fixed time horizon $T$, if $G(b,t)>1$, and $F(x)=\Theta(x^\alpha)$ with $\alpha>1$, then the expected regret of UCB-List $\mathbb{E}[R_T]$ is upper bounded by
\begin{equation*}
	\sum\limits_{a\neq a^*}\hspace{-4pt}\bigg[\cfrac{8\Delta_a\big(G(b,t)\hspace{-2pt}-\hspace{-2pt}1\big)\hspace{-2pt}+\hspace{-2pt}8\Delta_{max}}{\big(G(b,t)\hspace{-2pt}-\hspace{-2pt}1\big)\Delta_a^2} \ln T + 4\Delta_a+\cfrac{4\Delta_{max}}{G(b,t)\hspace{-2pt}-\hspace{-2pt}1}\bigg],
\end{equation*}
with the expected payment $\mathbb{E}[B_T]$ upper bounded by
\begin{equation*}
	\cfrac{2G(b,t)+1}{G(b,t)-1}\bigg[\cfrac{8b\ln T}{\Delta_{min}^2}+\sum_{a\neq a^*}\bigg(\cfrac{8b\ln T}{\Delta_a^2}+4b\bigg)\bigg].
\end{equation*}
\end{restatable}

\begin{remark}{\em
	Without any tuning parameter, the UCB-List policy adapts to a larger range of systems.
	The system parameters such as means of arms $\bm{\mu}$ or their gap summation $\sum_{a\neq a^*}\Delta_a$ play an important role in both regret and total payment.
	As $\sum_{a\neq a^*}\Delta_a$ decreases (implying it is harder to differentiate $a^*$), longer exploration and exploitation phases are needed, resulting in larger expected regret and total payment.
	Also, similar to Theorem~\ref{thm-etc}, as $G(b,t) \downarrow 1$, the expected regret and expected total payment are closer to being linear, because of the weak incentive effect.
	}
\end{remark}

\begin{proof}[Proof Sketch of Theorem~\ref{thm-ucb-list}]
We provide a proof sketch here and relegate the details to the supplementary material.
The expected time for initialization can be upper bounded by $O(1)$ trivially.
By the law of total expectation, we have:
\begin{align*}
	\mathbb{E}[R_T]\leq &\underbrace{\mathbb{E}[R_{\tau_1}]}_\textbf{(a)}+\underbrace{\mathbb{E}[R_{\tau_2}-R_{\tau_1}\mid \hat a^*=a^*]}_\textbf{(b)} \\
	&+\underbrace{\mathbb{E}[R_T-R_{\tau_2}\mid \hat a^*=a^*]}_\textbf{(c)}+\underbrace{T\cdot \mathbb{P}(\hat a^*\neq a^*)}_\textbf{(d)}.
\end{align*}
In what follows, we will bound the four terms on the right-hand-side one by one.

\textbf{(a)} In the exploration phase, since the regret results from the pulls of sub-optimal arms, the expected regret at time step $\tau_1$ can be written as $\mathbb{E}[R_{\tau_1}]=\sum_{a\neq a^*}\Delta_a\mathbb{E}[T_a(\tau_1)]$.
Thus, term \textbf{(a)} can be bounded if we upper bound $\mathbb{E}[T_a(\tau_1)]$ for each $a\in A$.
Let $U(t)$ denote the set of arms that can get payment at time $t$.
Consider the following two cases: \textbf{(i)} At time $t\leq\tau_1$, $a^*\in U(t)$ and there exists at least one suboptimal arm $a\in A, a\neq a^*$ such that $a\in U(t)$.
In this case we upper bound the probability $\mathbb{P}\big(\exists a\neq a^*: a\in U(t), a^*\in U(t)\big)$, and by using the Chernoff-Hoeffding bound, we obtain that when $T_a(t)\geq (8\ln T)/\Delta_a^2$ we have $\mathbb{P}\big(\exists a\neq a^*: a\in U(t), a^*\in U(t)\big)\leq 2T^{-1}$.
Thus, in this case, the expected regret is contributed by a suboptimal arm $a$ is $\Delta_a\mathbb{E}[T_a(t)]\leq (8\ln T)/\Delta_a + 2\Delta_a$;
\textbf{(ii)} At time $t\leq\tau_1$, $a^*$ is eliminated by some suboptimal arm $a\in U(t)$.
With the Chernoff-Hoeffding bound, we obtain $\mathbb{P}\big(\exists a\neq a^*: a\in U(t), a^*\notin U(t)\big)\leq 2T^{-1}$.
Summing over all possible cases and all suboptimal arms, $\mathbb{E}[R_{\tau_1}]$ is bounded by:
\begin{equation*}
	\mathbb{E}[R_{\tau_1}]
	\leq \sum_{a\neq a^*}\cfrac{8\ln T}{\Delta_a}+4\Delta_a.
\end{equation*}
\textbf{(b)} In the exploitation phase, the expected regret $\mathbb{E}[R_{\tau_2}-R_{\tau_1}\mid \hat a^*=a^*]$ is upper bounded by $O(\mathbb{E}[\tau_2-\tau_1])$ since
\begin{equation*}
	\mathbb{E}[R_{\tau_2} - R_{\tau_1} \mid\hat a^* = a^*]
	\leq \cfrac{\Delta_{max}}{G(b)+1}\cdot\mathbb{E}[\tau_2-\tau_1].
\end{equation*}
In term \textbf{(a)}, the upper bound of $\mathbb{E}[R_{\tau_1}]$ implies that each suboptimal arm $a$ is pulled at least $(8\ln T)/\Delta_a^2$ with $a^*$ being pulled at least $(8\ln T)/\Delta_{min}^2$ times, similar to the proof of Theorem~\ref{thm-etc} we obtain the upper bound of both $\mathbb{E}[\tau_1]$ and $\mathbb{E}[\tau_2-\tau_1]$.
This leads to the upper bounds of both $\mathbb{E}[R_{\tau_2}-R_{\tau_1}\mid \hat a^*=a^*]$ and $\mathbb{E}[B_T]= (\mathbb{E}[\tau_1]+\mathbb{E}[\tau_s-\tau_1])b$.

\textbf{(c)} This term represents the expected regret from $\tau_2$ to $T$.
Similar to the proof of Theorem~\ref{thm-etc}, this part of expected regret is bounded by $O\big((\log T)^{1-\gamma}e^{-(\log T)^\gamma}\big)$, $\gamma\in(0,1/4)$.

\textbf{(d)} The probability $\mathbb{P}(\hat a^*\neq a^*)$ can be bounded by $O(T^{-1})$ since $\mathbb{P}(\hat a^*\neq a^*)=\mathbb{P}\big(\exists a\neq a^*: a\in U(t), a^*\notin U(t)\big)$, which can be bounded by $2T^{-1}$ as in \textbf{(a)}-case \textbf{(ii)}.

Combining steps \textbf{(a)}--\textbf{(d)} yields the result stated in the theorem and the proof is complete.
\end{proof}

%% file: Sec5_Simulation/Simulation.tex

\section{Simulations}\label{sec:simulations}

In this section, we conduct simulations to evaluate the performances of AL$n$ETC and UCB-List policies.

\subsection{Comparisons with Baselines}
\label{baseline}
\begin{figure}[h!]
\centerline{
     \begin{subfigure}[b]{0.24\textwidth}
        	\centering
    	\includegraphics[trim=0 0 0 -15, width=\textwidth]{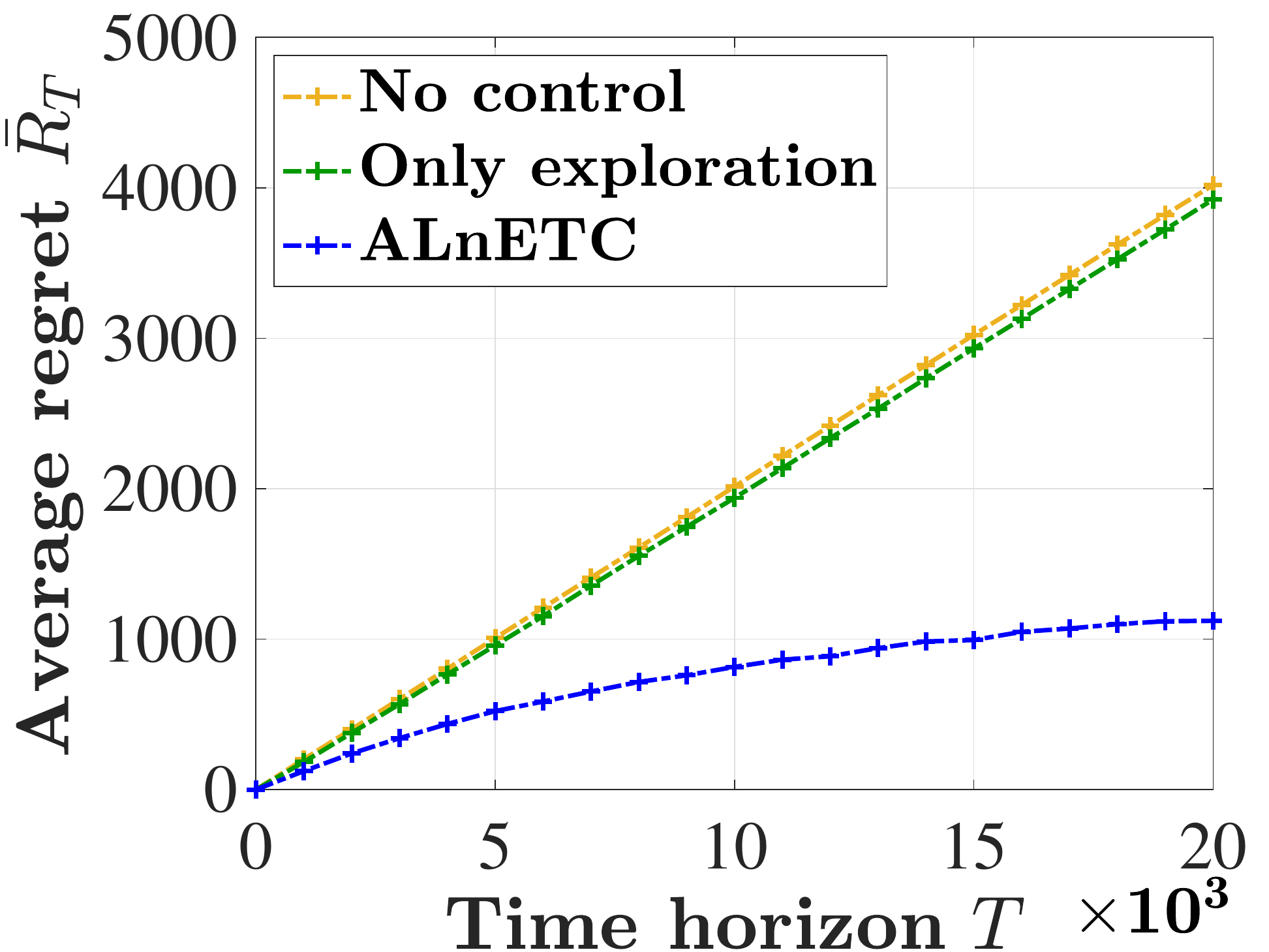}
    \end{subfigure}
    \hfill
    \begin{subfigure}[b]{0.24\textwidth}
        	\centering
    	\includegraphics[trim=0 0 0 -15, width=\textwidth]{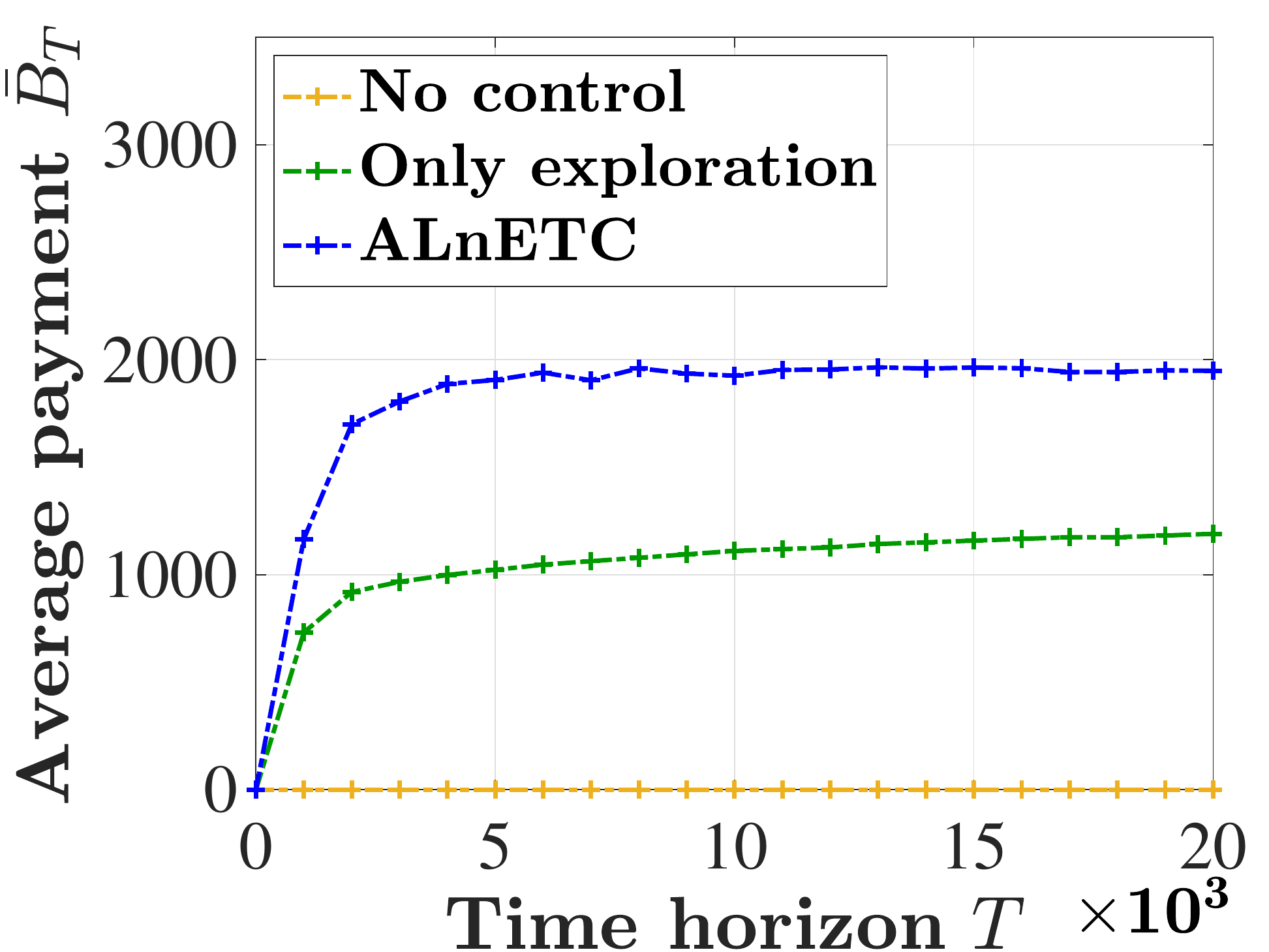}
    \end{subfigure}
    }
    \caption{Comparison of AL$n$ETC and baselines.}
    \label{fig1}
\end{figure}

We first compare the AL$n$ETC policy with two baselines: i) no incentive control, and ii) with incentive control only during exploration.
We only compare AL$n$ETC with the baselines since UCB-List outperforms AL$n$ETC (to be discussed next).
The simulation setting is as follows: a two-armed model with means $\bm{\mu} = [0.3, 0.5]$ and initial biases $\bm{\theta} = [100, 1]$, the feedback function $F(x)=x^\alpha$ with $\alpha = 1.5$ and payment $b=1.5$ with an incentive impact function $G(x,t) = x$.
We use the optimal AL$n$ETC parameter $q=15$.
The results are shown in Fig.~\ref{fig1}, where each data point is averaged over $1000$ trials.
We observe that the average regret under no incentives grows linearly due to the large initial bias toward the suboptimal arm and self-reinforcing preferences.
The average regret under partial incentive is also linear since the incentive is insufficient to offset the initial bias toward the suboptimal arm.
In contrast, the average regret of AL$n$ETC policy follows a $\log(T)$ growth rate.

\subsection{Comparisons with imperfect conditions}
\begin{figure}[h!]
\centerline{
     \begin{subfigure}[b]{0.24\textwidth}
        	\centering
    	\includegraphics[trim=0 0 0 0, width=\textwidth]{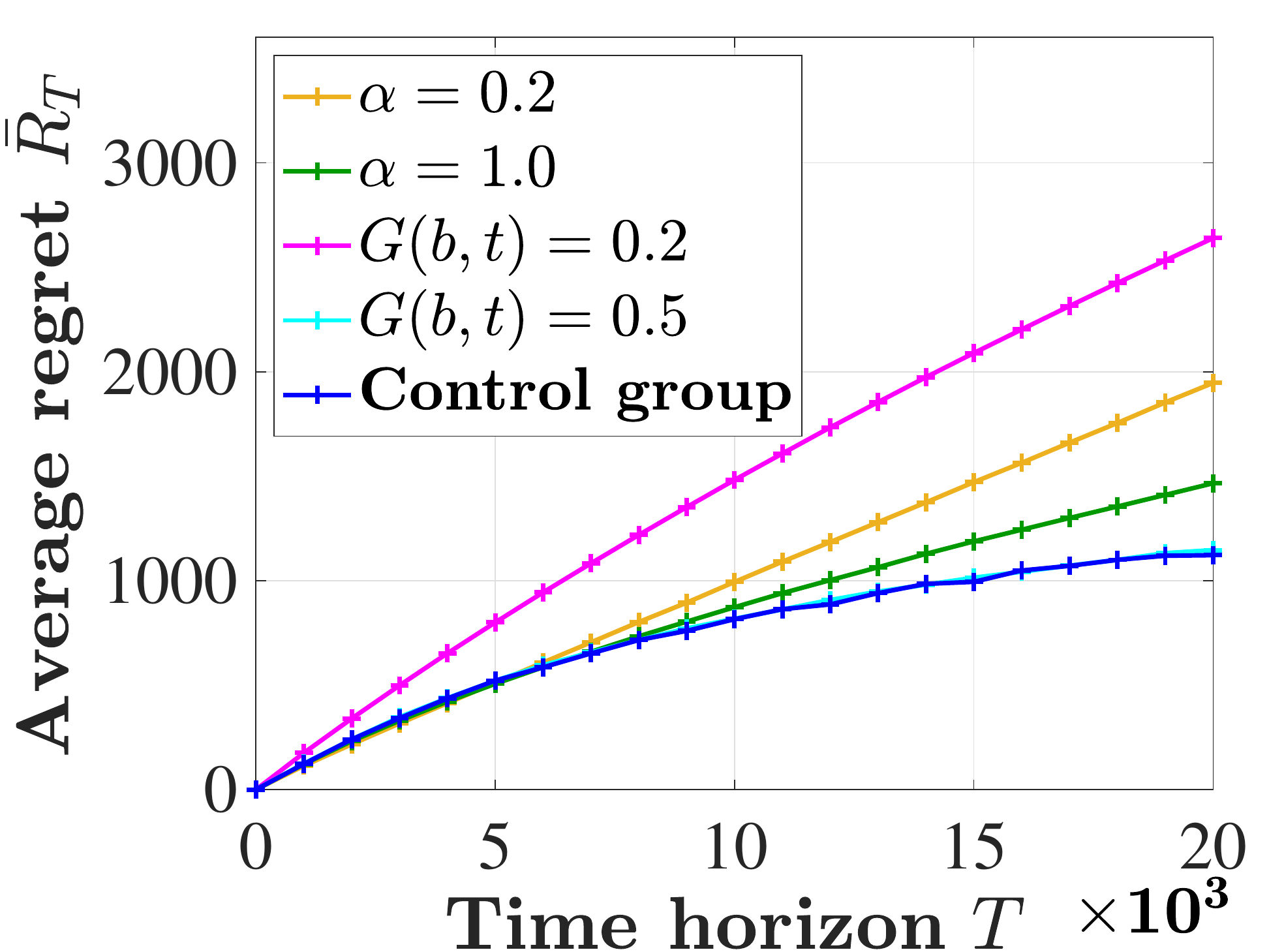}
   		\caption{Performance of AL$n$ETC.}
    \end{subfigure}
    \hfill
    \begin{subfigure}[b]{0.24\textwidth}
        	\centering
    	\includegraphics[trim=0 0 0 0, width=\textwidth]{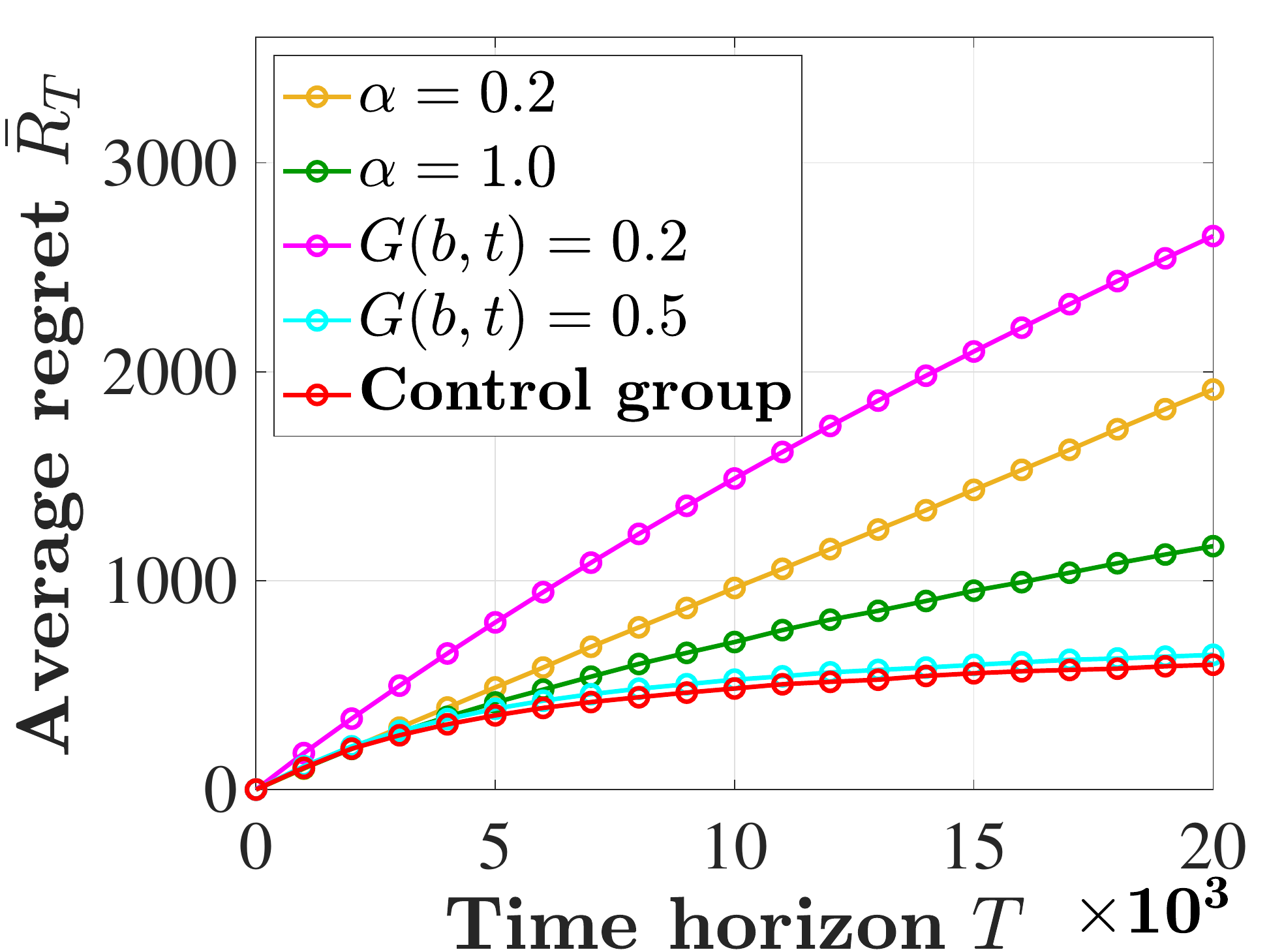}
   		\caption{Performance of UCB-List.}
    \end{subfigure}
    }
    \caption{Comparisons of imperfect conditions.}
    \label{assumptions}
\end{figure}

In real-world applications, some of our model conditions may not always hold (e.g., the conditions $G(b,t)>1$ and $F(x)=\Theta(x^\alpha)$ with $\alpha>1$).
Therefore, we conduct the simulations to study the robustness of our proposed policies.
The system setting in the group with incentive is almost the same as that in Section~\ref{baseline}: a two-armed model with means $\bm{\mu} = [0.3, 0.5]$ and initial biases $\bm{\theta} = [100, 1]$, the feedback function $F(x)=x^\alpha$.
The key difference is that, in this study, we set $\alpha\leq 1$ and $G(b,t)<1$ (i.e., the conditions in our theoretical results are not satisfied).
Specifically, we set the value of $G(b,t)$ to be $0.5$ and $0.2$, implying a weaker incentive impact.
Also, we choose the value of $\alpha$ to be $1.0$ and $0.2$, implying a weaker self-reinforcing preference strength.
We use the optimal AL$n$ETC parameter $q=15$.
The results are shown in Fig.~\ref{assumptions}, where each data point is averaged over $1000$ trials.
We observe that as the values of $\alpha$ and $G(b,t)$ decrease, the average regrets of both policies increase.
Specifically, when the incentive impact $G(b,t)$ becomes small enough, or the self-reinforcing preference strength is weak enough (e.g., $\alpha \leq 1$), the regrets of both policies no longer exhibit sub-linear trends.

\subsection{Comparisons between AL$n$ETC and UCB-List}
\begin{figure}[t!]
\centerline{
     \begin{subfigure}[b]{0.24\textwidth}
        	\centering
    	\includegraphics[trim=0 0 0 -15, width=\textwidth]{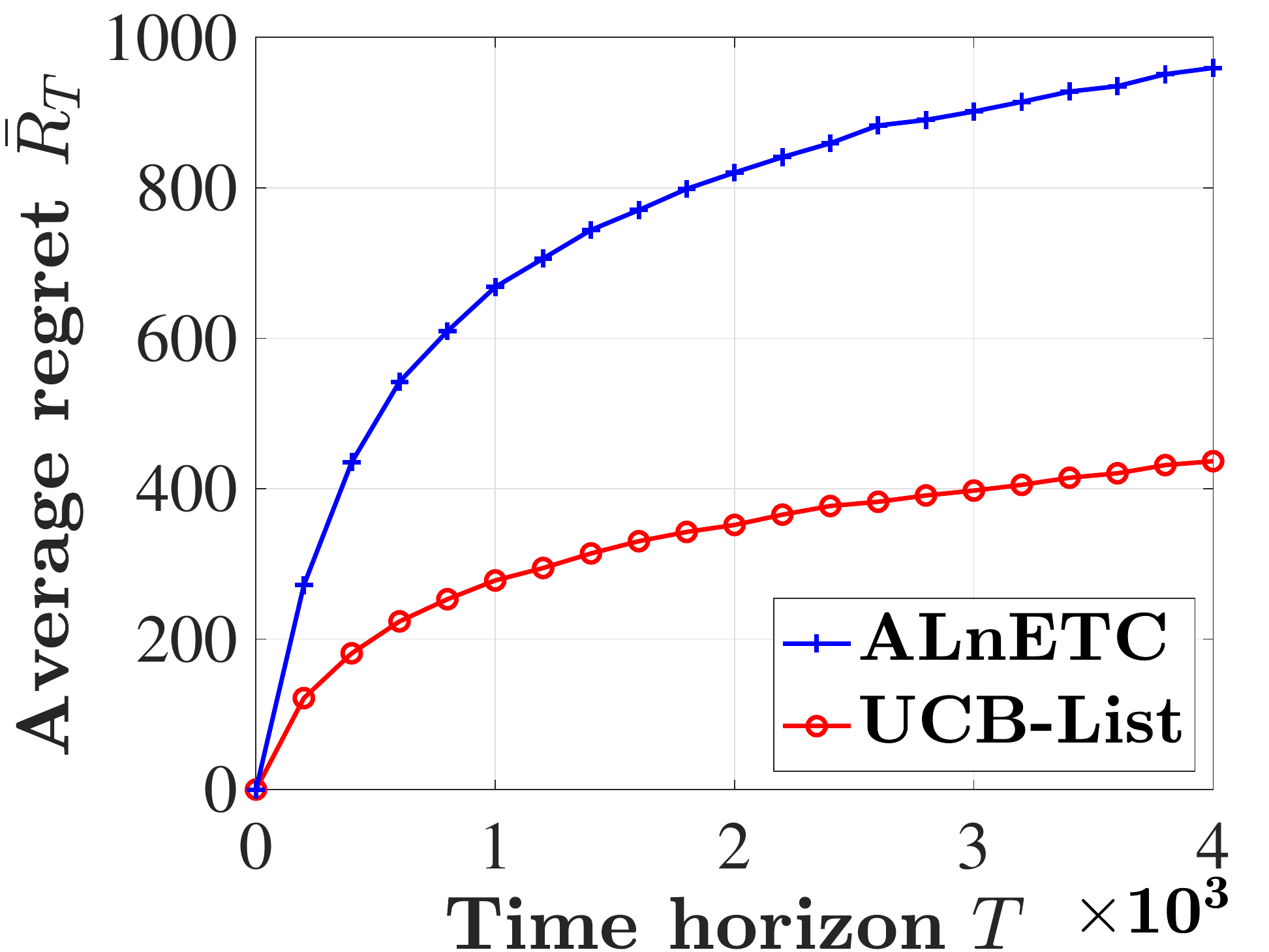}
    \end{subfigure}
    \hfill
    \begin{subfigure}[b]{0.24\textwidth}
        	\centering
    	\includegraphics[trim=0 0 0 -15, width=\textwidth]{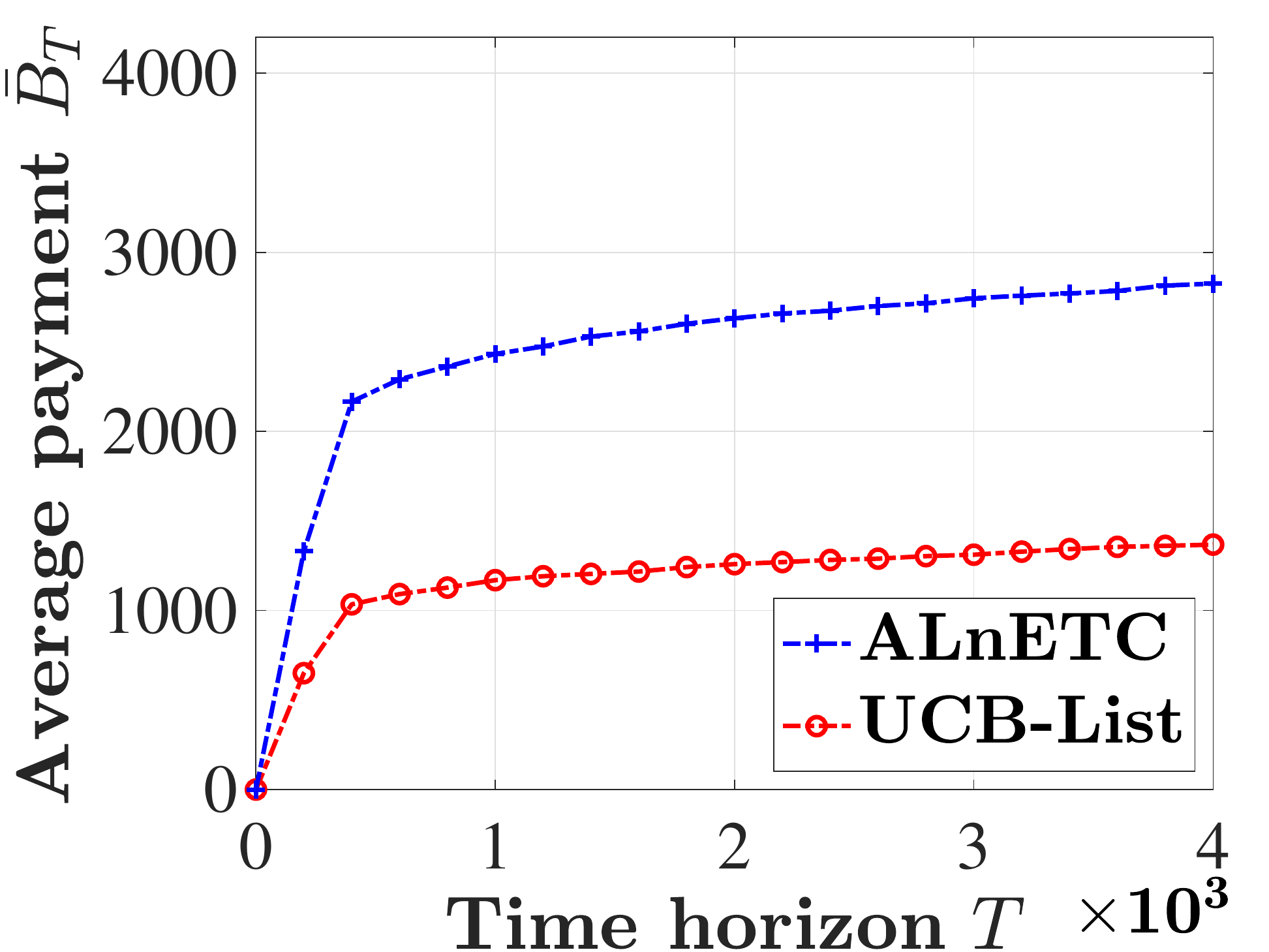}
    \end{subfigure}
    }
    \vspace{-2.5mm}
    \caption{Benchmark results.}
    \label{fig2}
\end{figure}

\begin{figure}[t!]
\centerline{
     \begin{subfigure}[b]{0.24\textwidth}
        	\centering
    	\includegraphics[trim=0 0 0 0, width=\textwidth]{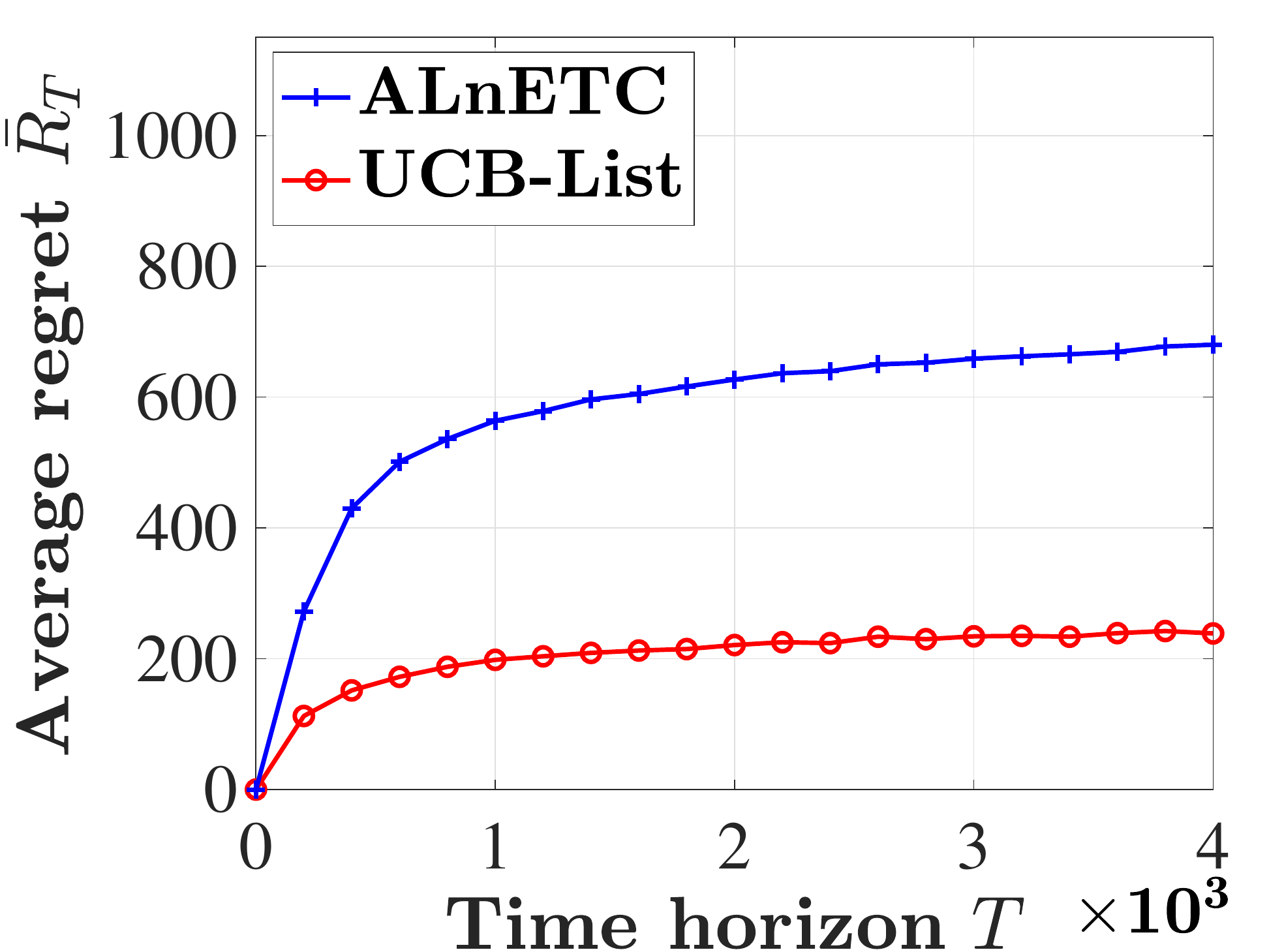}
    \end{subfigure}
    \hfill
    \begin{subfigure}[b]{0.24\textwidth}
        	\centering
    	\includegraphics[trim=0 0 0 0, width=\textwidth]{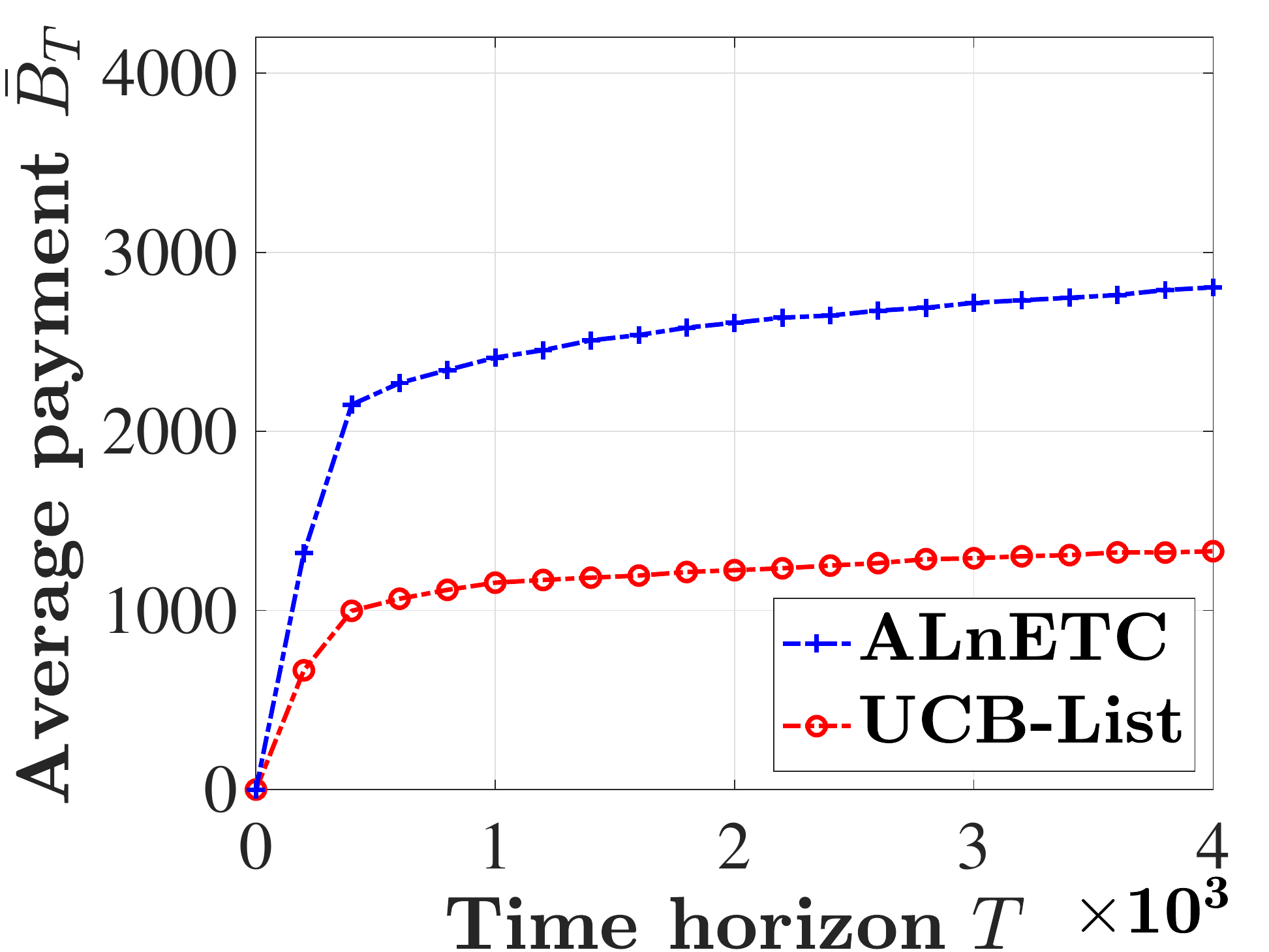}
    \end{subfigure}
    }
    \vspace{-2.5mm}
    \caption{Policy performance with parameter $\alpha = 2$.}
    \label{fig3}
\end{figure}

\begin{figure}[t!]
\centerline{
     \begin{subfigure}[b]{0.24\textwidth}
        	\centering
    	\includegraphics[trim=0 0 0 0, width=\textwidth]{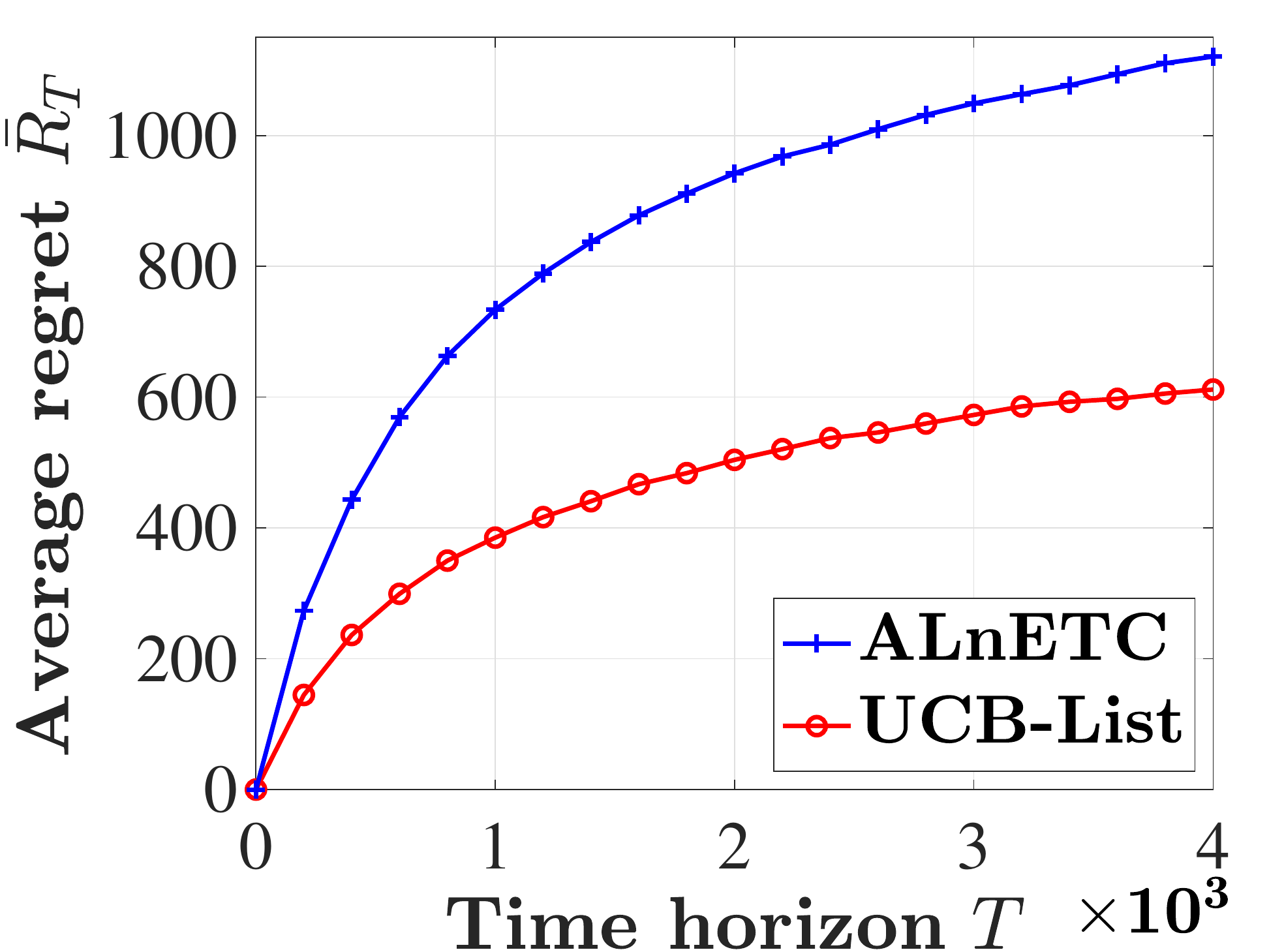}
    \end{subfigure}
    \hfill
    \begin{subfigure}[b]{0.24\textwidth}
        	\centering
    	\includegraphics[trim=0 0 0 0, width=\textwidth]{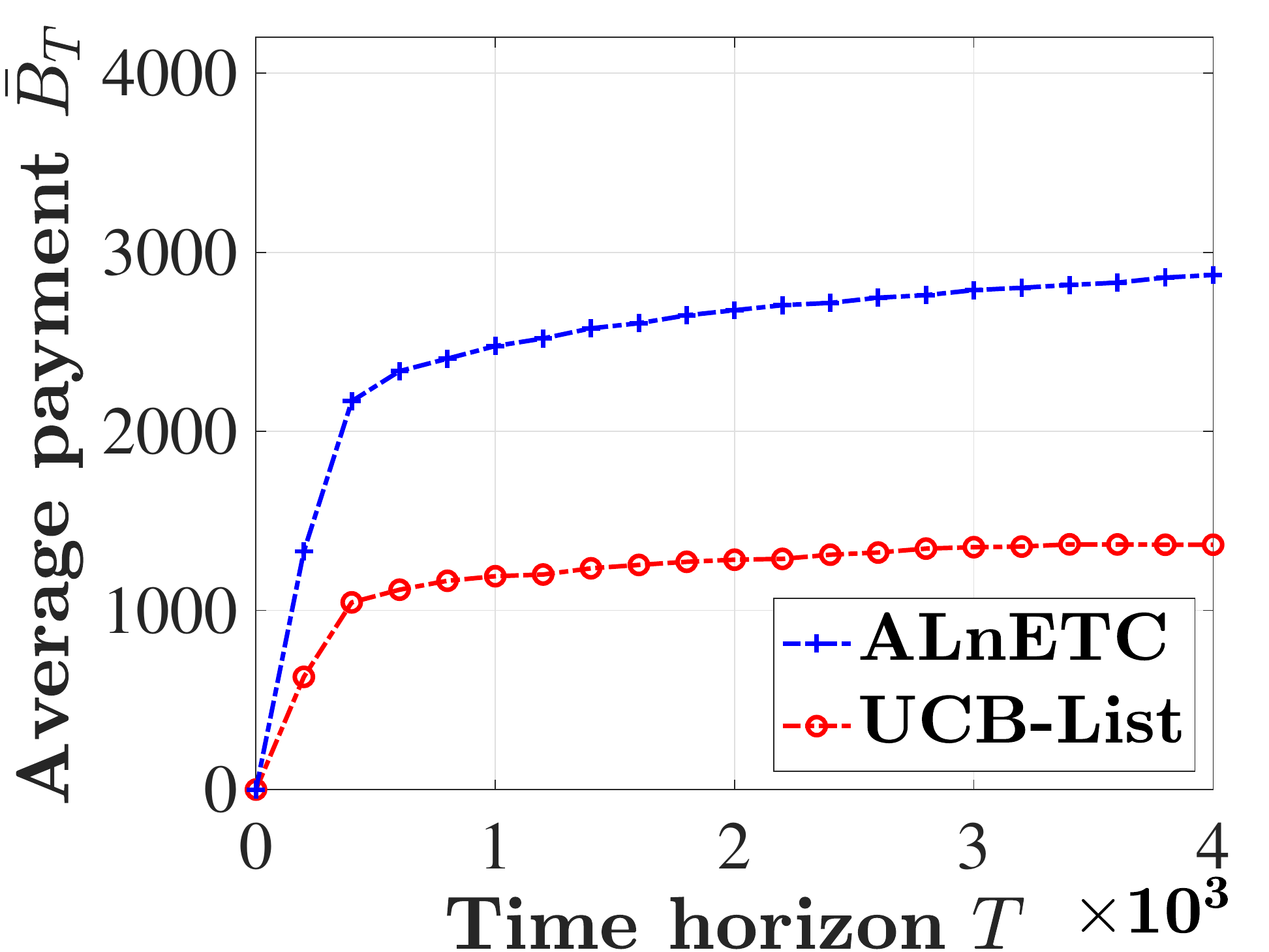}
    \end{subfigure}
    }
    \vspace{-2.5mm}
    \caption{Policy performance with parameter $\bm{\theta}=[50, 50, 1]$.}
    \label{fig4}
\end{figure}

\begin{figure}[t!]
\centerline{
     \begin{subfigure}[b]{0.24\textwidth}
        	\centering
    	\includegraphics[trim=0 0 0 -15, width=\textwidth]{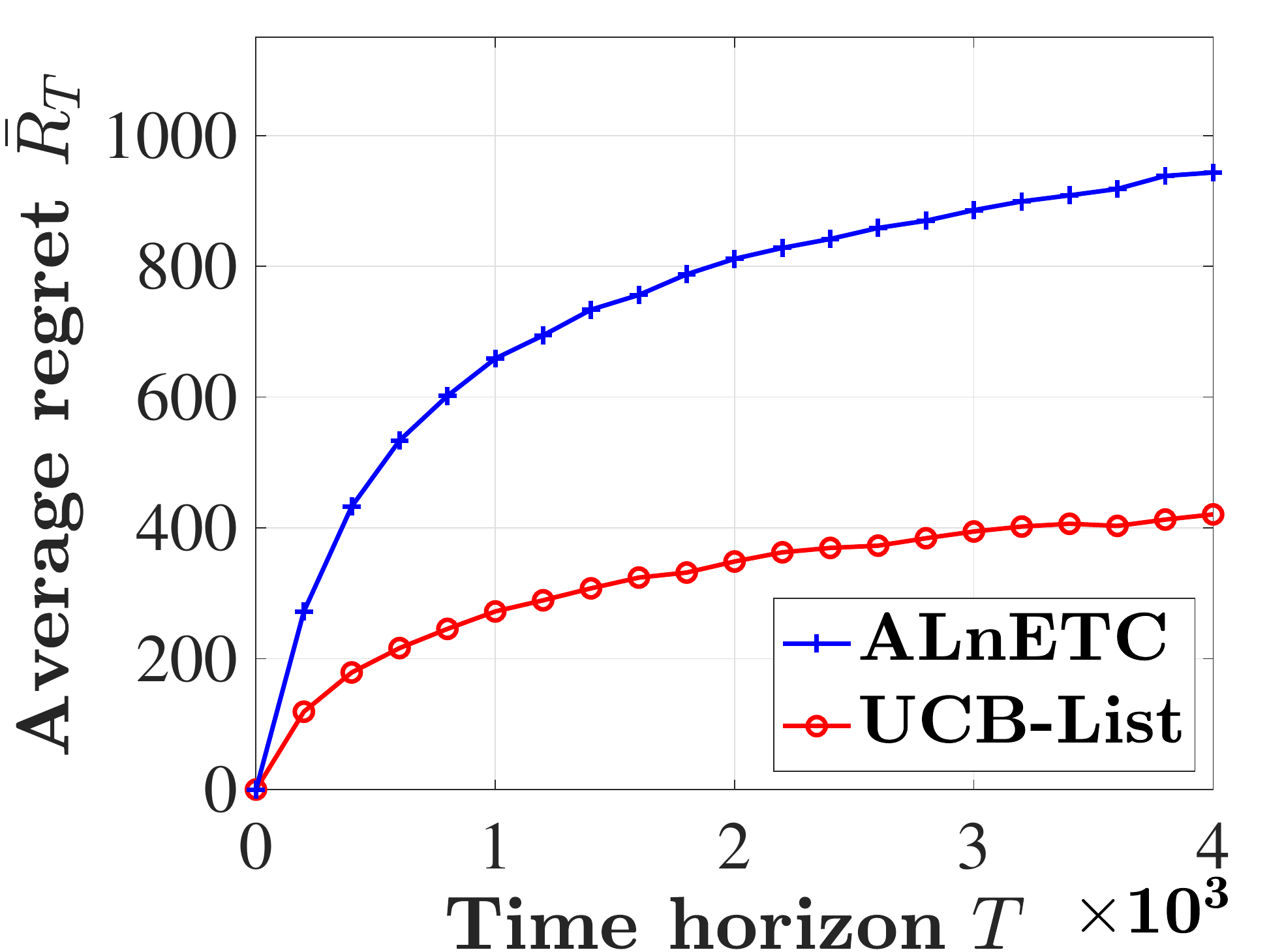}
    \end{subfigure}
    \hfill
    \begin{subfigure}[b]{0.24\textwidth}
        	\centering
    	\includegraphics[trim=0 0 0 -15, width=\textwidth]{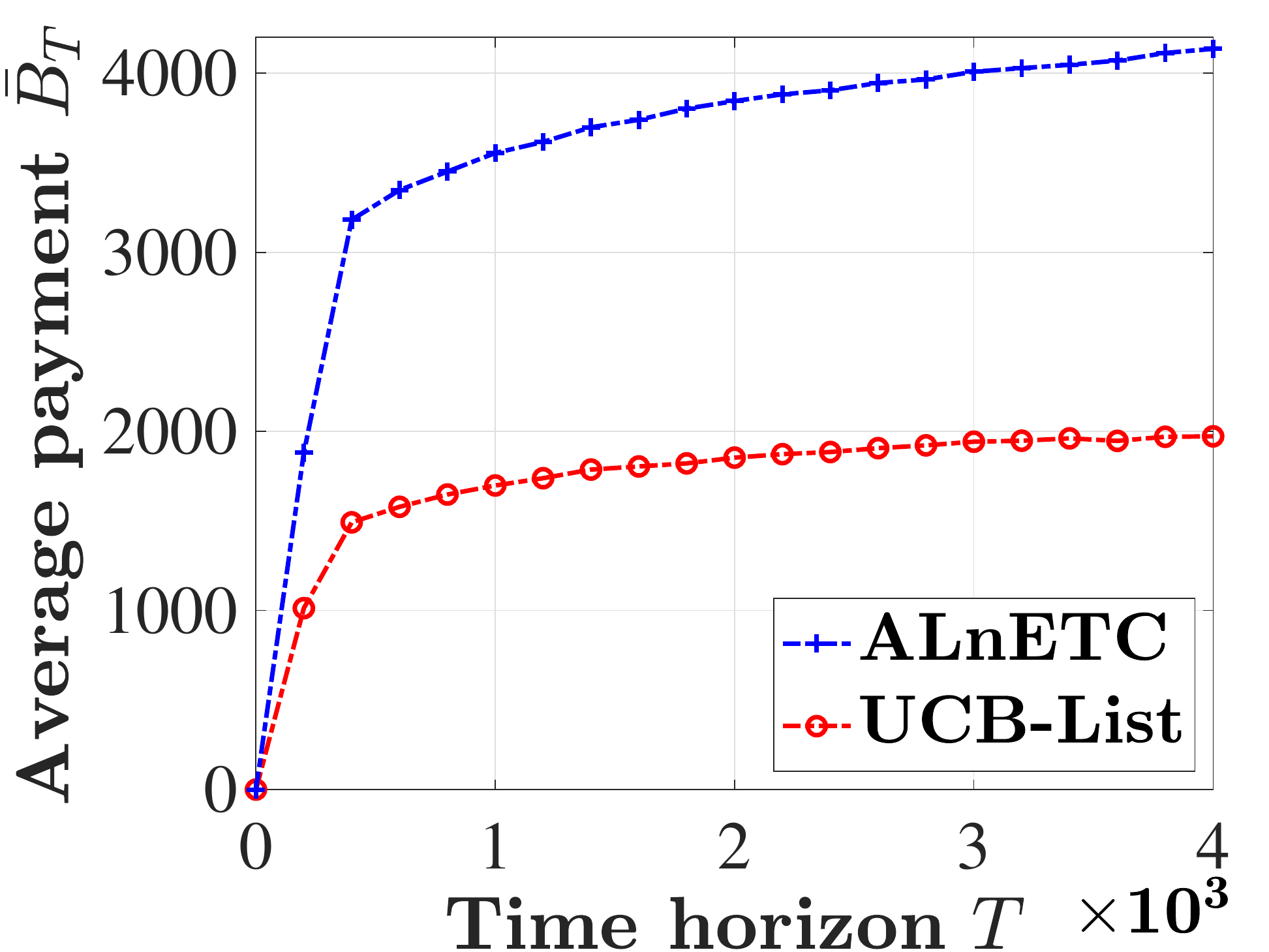}
    \end{subfigure}
    }
    \vspace{-2.5mm}
    \caption{Policy performance with parameter $b=1.8$.}
    \label{fig5}
\end{figure}

Finally, we compare AL$n$ETC and UCB-List.
The simulation setting is as follows: a three-armed model with means $\bm{\mu} = [0.2, 0.4, 0.6]$ and initial biases $\bm{\theta} = [10, 10, 1]$, the feedback function $F(x)=x^\alpha$, $\alpha = 1.5$ and payment $b=1.2$ with an incentive impact function $G(x,t) = x$.
For AL$n$ETC, we set the optimal parameter $q=20$.
Four groups of simulations are conducted and the results are shown in Fig.~\ref{fig2}-\ref{fig5}, where each data point is averaged over $1000$ trials.
Fig.~\ref{fig2} illustrates the performance of both average regret and total payment.
Fig.~\ref{fig2} also serves as a benchmark for comparisons with other three groups of results.
In each of Figs.~\ref{fig3}--\ref{fig5}, only one parameter is changed compared to the benchmark group.
This helps us observe the changes in average regret and total payment.
In Fig.~\ref{fig3}, all settings are the same as Fig.~\ref{fig2} except $\alpha = 2$.
In Fig.~\ref{fig4}, all settings are the same as those in Fig.~\ref{fig2} except $\bm{\theta} = [50, 50, 1]$.
In Fig.~\ref{fig5}, all settings are the same as Fig.~\ref{fig2} except $b = 1.8$.

The results show that both policies achieve $O(\log T)$ average regrets and $O(\log T)$ average total payment.
This indicates that: i) both policies balance the exploration-exploitation trade-off so that an order-optimal regret can be reached; 
ii) both policies balance the trade-off between maximizing the total reward and keeping the total payment growing at rate $O(\log T)$.
In Fig.~\ref{fig3}, the results show that both policies achieve a smaller average regret, because the self-reinforcing preferences are easier to converge to the incentivized arm under a larger $\alpha$.
Also, AL$n$ETC incurs a higher total payment because it incentivizes the pulling of sub-optimal arms more often.
In Fig.~\ref{fig4}, both policies have larger average regrets because it takes more effort for both policies to mitigate the larger initial biases.
In Fig.~\ref{fig5}, as the payment for each time step increases from $1.5$ to $1.8$, the average regrets are not affected significantly, while the total payments increases correspondingly.
Thus, a proper amount of payment depends on specific system parameters.

%% file: Sec6_Conclusion/Conclusion.tex

\section{Conclusion}\label{sec:conclusion}

We proposed and studied an incentivized bandit model with self-reinforcing preferences.
Two policies are proposed to achieve $O(\log T)$ expected regrets with $O(\log T)$ incentivized costs, under the condition that the feedback function satisfies $F(x)=\Theta(x^\alpha)$ for $\alpha>1$.
We conjecture that the feedback can be extended to a larger class of nonlinear functions.
We note that the area of incentivized MAB with self-reinforcing preferences remains under-explored.
Future works include, for example, the design of incentive schemes that can be time-varying in each time step, which can either depend on the current state, or be restricted by certain conditions.
The self-reinforcing preferences can also be viewed as contexts, and thus this setting can be modeled by leveraging the contextual bandit framework with more interesting properties.

\section*{Acknowledgements}
This work has been supported in part by NSF grants CAREER CNS-2110259, CCF-2110252, ONR grant N00014-17-1-2417, and a Google Faculty Research Award.

We thank the anonymous reviewers for their careful reading of our manuscript and their many insightful comments and suggestions.

%% file: Sec_Proof/Proof.tex
\appendix
\onecolumn

\clearpage

\begin{appendices}
{\bf\LARGE Supplementary Material}

\section{Proof of Lemma \ref{prop}}
\begin{figure}[h!]
    \centering
    \includegraphics[width=0.55\textwidth]{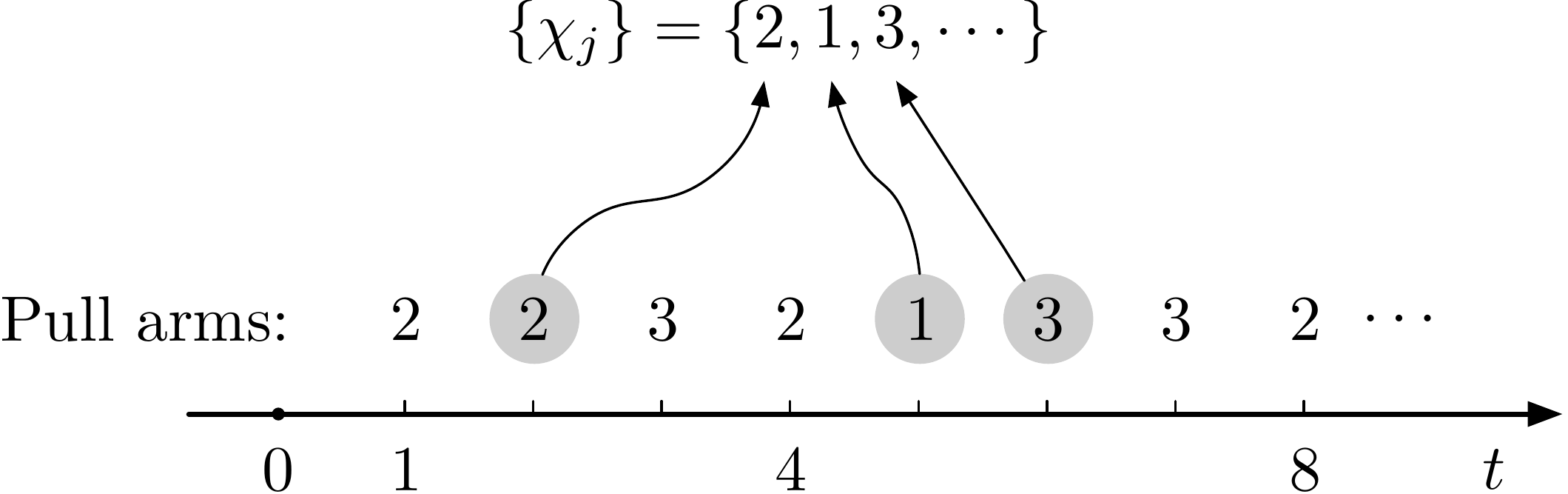}
    \caption{This figure shows an instance of sequence $\lbrace\chi_j\rbrace$. At time step $t=1$, arm $2$ is pulled and generates $0$ reward. At time step $t=2$, arm $2$ is pulled and generates a unit reward. Thus, the first element $\chi_1$ in $\lbrace\chi_j\rbrace$ is the arm index $2$ that generates the first unit reward. The subsequent elements in the sequence are generated similarly.}
    \label{fig:order}
\end{figure}

\RestateLEMMA*
Let the sequence $\lbrace\chi_j\rbrace_{j=1}^\infty$ be the arm order that generates a unit reward in our model without the participation of incentive, such that $\chi_j$ indicates the arm that generates the $j$-th unit reward, as shown in Figure~\ref{fig:order}.
Next we will construct a sequence that has the same conditional distribution as $\lbrace \chi_j\rbrace$.

Our main mathematical tool is the \textit{improved exponential embedding} method.
For each arm $i\in A$, we let $\lbrace r_i(n)\rbrace$ be a collection of independent exponential random variables such that $\mathbb{E}[r_i(n)]={1\over \mu_iF(n + \theta_i)}$.
We define set $B_i \coloneqq \lbrace \sum_{k=0}^n r_i(k)\rbrace_{n=0}^\infty$, where each element $\sum_{k=0}^n r_i(k)$ represents the random time needed for arm $i$ to get $n$ accumulative reward, and define set $G = B_1\cup B_2\cup \cdots \cup B_m$.
Let $\zeta_1$ be the smallest number in $G$ and in general let $\zeta_j$ be the $j$-th smallest number in $G$.
Next, we define a new random sequence $\lbrace \zeta_j \rbrace$, by making the $j$-th element of the sequence be the arm $i$ if $\zeta_j\in B_i$.
Then, we have the following lemma (to be proved later):

\begin{lemma}
\label{seq-construct}
Given the previous reward history $\mathcal{F}_{j-1}$, the constructed sequence $\lbrace \zeta_j\rbrace$ is equivalent in conditional distribution to the sequence $\lbrace\chi_j\rbrace$.
\end{lemma}
Next, we formally define the notion of attraction time.
\begin{definition}[Attraction time]
Let $N$ denote the attraction time, such that after this time step $N$, monopoly happens, i.e., only one arm has positive probability to generate rewards.
\end{definition}

\noindent Necessity: if $\alpha > 1$ then $\mathbb{P}(N < \infty) = 1$.
With the help of improved exponential embedding, the time until the accumulative reward of arm $i\in A$ approaches infinity is $\sum_{k=0}^\infty r_i(k)$.
If the condition $\sum_i {1\over F(i)}<\infty$ is satisfied, then we have
\begin{equation*}
    \mathbb{E}\big[\sum_{k=0}^\infty r_i(k)\big] = {1\over \mu_i}\sum_{k=0}^\infty {1\over F(k+\theta_i)}<\infty.
\end{equation*}
So for each arm $i\in A$, $\mathbb{P}(\sum_{k=0}^\infty r_i(k)<\infty)=1$.
Let $a=\arg\min_{i\in A}\lbrace \sum_{k=0}^\infty r_i(k)\rbrace$, then for each $b\neq a$, there exists a finite number $K_b$ such that
\begin{equation*}
    \sum_{k=0}^{K_b} r_b(k)< \sum_{k=0}^\infty r_a(k) < \sum_{k=0}^{K_b+1} r_b(k).
\end{equation*}
Thus if we let $N:= \max_{i\in A, i\neq a}\lbrace \sum_{k=0}^{f_i(k)} r_i(k)\rbrace$, then after this time $N$, only arm $a$ can generate rewards.\\
\\
Sufficiency: if $\mathbb{P}(N < \infty) = 1$ then $\sum_i {1\over F(i)}<\infty$.
If we show that when $\sum_i {1\over F(i)}=\infty$ we have $\mathbb{P}(N = \infty) > 0$, then the proof is done.
When $\sum_i {1\over F(i)}=\infty$, we have
\begin{equation*}
    \mathbb{E}\big[\sum_{k=0}^\infty r_i(k)\big] = {1\over \mu_i}\sum_{k=0}^\infty {1\over F(k+\theta_i)}\rightarrow\infty.
\end{equation*}
Thus for any $i\in A$ it takes infinite time to accumulate infinite reward, which implies $\mathbb{P}(N = \infty) > 0$.
In fact, in this case $\mathbb{P}(N = \infty)=1$.
We refer readers to \citet{khanin2001probabilistic} and \citet{oliveira2009onset} for further details.

\subsection{Proof of Lemma \ref{seq-construct}}
The proof of this lemma relies on the memoryless property of the exponential distribution as well as the following two facts:
\begin{fact}
If $X_1, \cdots, X_m (m\geq 2)$ are independent exponential random variables with parameter $\lambda_1, \cdots, \lambda_m$, respectively, then $\min(X_1, \cdots, X_m)$ is also exponential with parameter $\lambda_1 +\cdots + \lambda_m$.
\end{fact}
\begin{fact}
For two independent exponential random variables $X_1\sim exp(\lambda_1)$ and $X_2\sim exp(\lambda_2)$, $\mathbb{P}(X_1<X_2)={\lambda_1\over \lambda_1 + \lambda_2}$.
\end{fact}
Initially, in the sequence $\lbrace \zeta_j\rbrace$ when $j=1$, since the initial value for arm $i$ is its bias $\theta_i$, using the above two facts:
\begin{align*}
    \mathbb{P}(\zeta_1=i\mid\mathcal{F}_0)
    &= \mathbb{P}\bigg(r_i(0)<\min_{j\neq i}\lbrace r_j(0)\rbrace\bigg|\mathcal{F}_0\bigg)\\
    &= {\mu_iF(\theta_i)\over \sum_{j\in A}\mu_jF(\theta_j)}.
\end{align*}
In our model, each arm $i$ has probability $\mu_i\cdot \lambda_i(t)={{\mu_i}F(\theta_i) \over \sum_{j\in A}F(\theta_j)}$ to generate the first reward every time step before it does.
The value of element $\chi_1$ is a random variable following multinomial distribution with single trial, i.e., with $\mathcal{F}_0$, the event $\lbrace\chi_1=i\rbrace$ happens with probability $\mathbb{P}(\chi_1=i\mid\mathcal{F}_0)={{\mu_i}F(\theta_i) \over \sum_{j\in A}\mu_jF(\theta_j)}$, and $\sum_{i\in A}\mathbb{P}(\chi_1=i\mid\mathcal{F}_0)=1$.
Thus
\begin{align*}
	\mathbb{P}(\zeta_1=i\mid\mathcal{F}_0) = \mathbb{P}(\chi_1=i\mid\mathcal{F}_0)
\end{align*}
Now suppose that before $\zeta_n$, each arm $a$ has been added to $N_a$. Then
\begin{align*}
    \mathbb{P}(\zeta_n=i\mid\mathcal{F}_{\zeta_{n-1}})
    &= \mathbb{P}\bigg(r_i(N_i+1)<\min_{j\neq i}\lbrace r_j(N_j+1)\rbrace\bigg|\mathcal{F}_{
    \zeta_{n-1}}\bigg)\\
    &= {\mu_iF(N_i + \theta_i)\over \sum_{j\in A}\mu_jF(N_j + \theta_j)}.
\end{align*}

Correspondingly in our model, each arm $i$ has probability $\mu_i\cdot \lambda_i(t)={{\mu_i}F(N_i + \theta_i) \over \sum_{j\in A}F(N_j + \theta_j)}$ to generate the next reward every time step before it does.
The value of element $\chi_n$ is a random variable following multinomial distribution with single trial, i.e., with $\mathcal{F}_{\chi_{n-1}}$, the event $\lbrace\chi_n=i\rbrace$ happens with probability $\mathbb{P}(\chi_n=i\mid\mathcal{F}_{\chi_{n-1}})={{\mu_i}F(N_i + \theta_i) \over \sum_{j\in A}\mu_jF(N_j + \theta_j)}$, and $\sum_{i\in A}\mathbb{P}(\chi_n=i\mid\mathcal{F}_{\chi_{n-1}})=1$.
Thus,
\begin{align*}
	\mathbb{P}(\zeta_n=i\mid\mathcal{F}_{\zeta_{n-1}}) = \mathbb{P}(\chi_n=i\mid\mathcal{F}_{\chi_{n-1}}).
\end{align*}

\section{Proof of Lemma \ref{dominance}}
\label{proof of dominance}
\RestateLEMMAD*
Recall that the definition of dominance is at time $t\geq\tau_n$, $S_{\hat a^*}(t)\geq\sum_{a\neq\hat a^*}S_a(t)$.
Thus arm $\hat a^*$ is expected to dominate at time $t\geq \tau_n$ if
\begin{equation*}
	\mu_{\hat a^*}\mathbb{E}[T_{\hat a^*}(t)] \geq \sum_{a\neq \hat a^*}\mu_a\mathbb{E}[T_a(t)].
\end{equation*}
We tighten this condition by narrowing the left-hand-side and amplifying the right-hand-side as follows:
\begin{align}
	&\mu_{\hat a^*}\mathbb{E}[T_{\hat a^*}(t)] \geq \sum_{a\neq \hat a^*}\mu_a\mathbb{E}[T_a(t)]\nonumber\\
	&\Rightarrow T_{\hat a^*}(\tau_n)+\mu_{\hat a^*}\mathbb{E}[T_{\hat a^*}(t)-T_{\hat a^*}(\tau_n)] \geq \sum_{a\neq \hat a^*}T_a(\tau_n)+\sum_{a\neq \hat a^*}\mu_a\mathbb{E}[T_a(t)-T_a(\tau_n)]\nonumber\\
	&\Rightarrow n+\mu_{\hat a^*}\mathbb{E}[T_{\hat a^*}(t)-T_{\hat a^*}(\tau_n)] \overset{\text{(i)}}{\geq} (\mu_{\hat a^*}\mathbb{E}[\tau_n] - n)+\sum_{a\neq \hat a^*}\mu_a\mathbb{E}[T_a(t)-T_a(\tau_n)]\nonumber\\
	&\Rightarrow n+\mu_{\hat a^*}\cfrac{G(b,t)}{G(b,t)+1}\mathbb{E}[t-\tau_n] \overset{\text{(ii)}}{\geq} (\mu_{\hat a^*}\mathbb{E}[\tau_n] - n)+\mu_{\hat a^*}\cfrac{\mathbb{E}[t-\tau_n]}{G(b,t)+1}\nonumber\\
	&\Rightarrow \mathbb{E}[t-\tau_n]\overset{\text{(iii)}}{\geq}\cfrac{\big(\mathbb{E}[\tau_n]-\cfrac{2n}{\mu_{\hat a^*}}\big)\big(G(b,t)+1\big)}{G(b,t)-1},\label{15}
\end{align}
where (i) is because arm $\hat a^*$ is pulled at least $n$ times during the exploration phase, (ii) is because by incentivizing arm $\hat a^*$, we have $\hat \lambda_{\hat a^*}(t)\geq{G(b,t)\over G(b,t)+1}$ and $\hat\lambda_a(t)\leq{1\over G(b,t)+1}$ for $a\neq \hat a^*$, and (iii) is the rearrangement.
Then we obtain the sufficient condition of dominance (\ref{15}).
Since time $\tau_s$ is defined as the earliest time to reach dominance, we can upper bound $\mathbb{E}[\tau_s-\tau_n]$ by
\begin{equation}
	\mathbb{E}[\tau_s-\tau_n]\leq\cfrac{\big(\mathbb{E}[\tau_n]-\cfrac{2n}{\mu_{\hat a^*}}\big)\big(G(b,t)+1\big)}{G(b,t)-1}.\label{t2}
\end{equation}
Next, we prove the following result for $\mathbb{E}[\tau_n]$.
\begin{lemma}
\label{tau-n bound}
	In AL$n$ETC, the expected exploration phase duration $\mathbb{E}[\tau_n]$ is upper bounded by $O(\log T)$.
\end{lemma}

\subsection{Proof of Lemma~\ref{tau-n bound}}
In AL$n$ETC, during the exploration phase at time step $t$, the agent offers payment $b$ to the user pulling arm $i$.
The probability that the arm $i$ generates reward is ${\lambda_i(t)+G(b,t)\over 1+G(b,t)}\cdot \mu_i>{G(b,t)\mu_i\over 1+G(b,t)}$.
Thus, the number of attempts for arm $i$ to generate a unit reward is a geometric random variable with parameter larger than ${G(b,t)\mu_i\over 1+G(b,t)}$.
By the policy, during the exploration phase, each arm generates at least $n$ accumulative reward. Then we obtain
\begin{equation}
    \mathbb{E}[\tau_n] \leq n\cdot \sum_{i\in A}{1+G(b,t)\over G(b,t)\mu_i} = O(n) = O(\log T).\label{t1}
\end{equation}

Lastly, it follows from Lemma~\ref{tau-n bound} that $\mathbb{E}[\tau_s]=\mathbb{E}[\tau_n] + \mathbb{E}[\tau_s-\tau_n]=O(\log T)$.
This completes the proof.

\section{Proof of Theorem \ref{thm-etc}}
\RestateETC*
In the rest of the proofs, for simplicity we will use the notations $\Delta_a = \mu^*-\mu_a$, $\mu_{min}=\min\limits_{a\in A} \mu_a$, $\Delta_{max}=\max\limits_{a\in A}\Delta_a$ and $\Delta_{min}=\min\limits_{a\in A}\Delta_a$.

By the law of total expectation, the expected regret up to $T$ is as follows:
\begin{align*}
    \mathbb{E}[R_T] &= \mathbb{E}[R_T \mid\hat a^* = a^*]\mathbb{P}(\hat a^* = a^*) + \mathbb{E}[R_T\mid\hat a^*\neq a^*]\mathbb{P}(\hat a^*\neq a^*)\\
    &\leq \mathbb{E}[R_T \mid\hat a^* = a^*] + T\cdot\mathbb{P}(\hat a^*\neq a^*).
\end{align*}
We want to bound both $\mathbb{E}[R_T \mid\hat a^* = a^*]$ and $\mathbb{P}(\hat a^*\neq a^*)$ to get the regret bound.
First we analyze the upper bound of the part $\mathbb{P}(\hat a^*\neq a^*)$.
We start with the following lemma.
\begin{lemma}
\label{best-arm-rate-greedy}
For each arm $a\neq a^*$, there exists a constant $\epsilon_a > 0$ independent of $n$ such that the following hold:
\begin{equation*}
    \mathbb{P}\bigg( \hat\mu_a(\tau_n) > \mu_a + {\Delta_a \over 2}\bigg)\leq 2e^{-2\epsilon_a n},
\end{equation*}
and
\begin{equation*}
    \mathbb{P}\bigg( \hat\mu_{a^*}(\tau_n) < \mu_{a^*} - {\Delta_a \over 2}\bigg)\leq 2e^{-2\epsilon_a n}.
\end{equation*}
\end{lemma}


\noindent Let arm $a=\arg\max_{i\in A, i\neq a^*}\hat\mu_i(\tau_n)$ denote the arm with largest sample mean and not equal to arm $a^*$ at time step $\tau_n$.
We have:
\begin{align*}
    \mathbb{P}(\hat a^*\neq a^*)
    &\leq \mathbb{P}\bigg(\hat\mu_a(\tau_n)\geq \hat\mu_{a^*}(\tau_n)\bigg)\\
    &\overset{\text{(i)}}{\leq} \mathbb{P}\bigg(\hat\mu_a(\tau_n)\geq \mu_a + {\Delta_a\over 2}\bigg) + \mathbb{P}\bigg(\hat\mu_{a^*}(\tau_n)\leq\mu_{a^*}-{\Delta_a \over 2}\bigg)\\
    &\overset{\text{(ii)}}{\leq} 4e^{-{n\Delta_a^2\over 2\mu_a}},
\end{align*}
where (i) is because $\mu_a+\Delta_a/2=\mu_{a^*}-\Delta_a/2$, and the event $\lbrace\hat\mu_a(\tau_n)\geq \hat\mu_{a^*}(\tau_n)\rbrace$ implies either $\lbrace\hat\mu_a(\tau_n)\geq \mu_a+\Delta_a/2\rbrace$ or $\lbrace\hat\mu_{a^*}(\tau_n)\leq \mu_{a^*}-\Delta_a/2\rbrace$, and (ii) follows by leveraging Lemma~\ref{best-arm-rate-greedy}.
Recall that, in the policy, we define $n=q\log T$. Thus, if $q\geq {2\max_{a\neq a^*}\mu_a\over\Delta_{min}^2}$, it then follows that $\mathbb{P}(\hat a^*\neq a^*)=O({1\over T})$.

Next, we analyze the upper bound of the part $\mathbb{E}[R_T \mid\hat a^* = a^*]$.
Let $\Gamma_t$ denote the accumulative reward up to time step $t$.
Then, we have:
\begin{align}
    \mathbb{E}[R_T \mid\hat a^* = a^*]
    &= \mathbb{E}[\Gamma^*_T] - \mathbb{E}[\Gamma_T\mid\hat a^* = a^*]\nonumber \\
    &= \mu^*\cdot T - \mathbb{E}[\Gamma_T\mid\hat a^* = a^*]\nonumber \\
    &= \mu^*\cdot T - \big( \mathbb{E}[\Gamma_{\tau_s}\mid\hat a^* = a^*] + \mathbb{E}[\Gamma_T-\Gamma_{\tau_s}\mid\hat a^* = a^*] \big ).\label{1}
\end{align}
During the exploration phase, since each arm generates rewards at least $n$ times, we obtain:
\begin{align}
    \mathbb{E}[\Gamma_{\tau_n}\mid\tau_n]
    &=\mathbb{E}\bigg[\sum_{i\in A}\big(n + (S_i(\tau_n) - n)\big)\bigg]\nonumber\\
    &= m\cdot n + \mathbb{E}\bigg[\sum_{i\in A}\big(T_i(\tau_n)\cdot\mu_i - n\big) \bigg]\nonumber\\
    &= m\cdot n + \sum_{i\in A}\mu_i\bigg(\mathbb{E}[T_i(\tau_n)]-{n\over\mu_i}\bigg)\nonumber\\
    &\geq m\cdot n + \mu_{min}\cdot\sum_{i\in A}\bigg(\mathbb{E}[T_i(\tau_n)]-{n\over\mu_i}\bigg)\nonumber\\
    &= m\cdot n + \bigg(\tau_n\cdot\mu_{min} - \mu_{min}\cdot\sum_{i\in A}{n\over \mu_i} \bigg)\nonumber\\
    &= \tau_n\cdot \mu_{min} + n\cdot \sum_{i\in A}{\mu_i-\mu_{min}\over \mu_i}.\label{2}
\end{align}
For each arm $a\in A$, let $L_a={F(q\ln T+\theta_a)\over\sum_{i\in A}F(\mu^*T+\theta_i)}$.
Thus at time $t\in\lbrace\tau_n+1,\ldots, T\rbrace$, we have
\begin{equation*}
	\mathbb{E}[\lambda_a(t)]=\mathbb{E}\bigg[\cfrac{F(S_a(t-1)+\theta_a)}{\sum_{i\in A}F(S_i(t-1)+\theta_i)}\bigg]\overset{\text{(i)}}{\geq}\cfrac{F(q\ln T+\theta_a)}{\sum_{i\in A}F(\mu^*T+\theta_i)}= L_a,
\end{equation*}
where (i) is obtained since at time $t> \tau_n$, $S_a(t-1)\geq q\ln T$ and $S_a(t-1)\leq \mu^*T$ for any $a\neq a^*$.

During the exploitation phase, the agent offers payment to users pulling arm $\hat a^*$, so using the bound in (\ref{2}) we obtain:
\begin{align}
    &\hspace{4.5mm}\mathbb{E}[\Gamma_{\tau_s}\mid\hat a^*=a^*, \tau_n, \tau_s]\nonumber\\
    &= \mathbb{E}[\Gamma_{\tau_n}\mid\tau_n] + \sum_{t=\tau_n+1}^{\tau_s}\mathbb{E}\bigg[{\lambda_{a^*}(t)+G(b,t)\over 1+G(b,t)}\cdot\mu^* + \sum_{i\in A}{\lambda_i(t)\over 1+G(b,t)}\cdot \mu_i \bigg] \nonumber \\
    &\geq \mathbb{E}[\Gamma_{\tau_n}\mid\tau_n] + \sum_{t=\tau_n+1}^{\tau_s}\mathbb{E}\bigg[{\lambda_{a^*}(t)+G(b,t)\over 1+G(b,t)}\cdot\mu^* + {(1-\lambda_{a^*}(t))\over 1+G(b,t)}\cdot \mu_{min}\bigg]\nonumber \\
    &= \mathbb{E}[\Gamma_{\tau_n}\mid\tau_n] + \sum_{t=\tau_n+1}^{\tau_s}\mathbb{E}\bigg[{G(b,t)\over 1+G(b,t)}\cdot\mu^* + {\mu_{min}\over 1+G(b,t)} + {\lambda_{a^*}(t)\Delta_{max}\over 1+G(b,t)}\bigg]\nonumber\\
    &\geq \mathbb{E}[\Gamma_{\tau_n}\mid\tau_n] + {\mu^*(\tau_s - \tau_n)G(b,t) \over 1 + G(b,t)} + {(\tau_s - \tau_n)\mu_{min} \over 1 + G(b,t)} + \cfrac{(\tau_s-\tau_n)L_{a^*}\Delta_{max}}{1+G(b,t)}\nonumber \\
    &\overset{\text{(i)}}{\geq} \tau_n\cdot \mu_{min} + n\cdot \sum_{i\in A}{\mu_i-\mu_{min}\over \mu_i} + {\mu^*(\tau_s - \tau_n)G(b,t) \over 1 + G(b,t)} + {(\tau_s - \tau_n)\mu_{min} \over 1 + G(b,t)} + \cfrac{(\tau_s-\tau_n)L_{a^*}\Delta_{max}}{1+G(b,t)}\nonumber\\
    &= n\sum_{i\in A}{\mu_i-\mu_{min}\over \mu_i} + {\mu^*G(b,t) + \mu_{min}+L_{a^*}\Delta_{max}\over 1+G(b,t)}\tau_s + \tau_n\cdot\mu_{min} - {\mu^*G(b,t)+\mu_{min} + L_{a^*}\Delta_{max}\over 1+G(b,t)}\tau_n \nonumber\\
    &= n\sum_{i\in A}{\mu_i-\mu_{min}\over \mu_i} + {\mu^*G(b,t) + \mu_{min}+L_{a^*}\Delta_{max}\over 1+G(b,t)}\tau_s - {\big(G(b,t)+L_{a^*}\big)\Delta_{max}\over 1+G(b,t)}\tau_n,\label{3}
\end{align}
where (i) is obtained by replacing $\mathbb{E}[\Gamma_{\tau_n}\mid\tau_n]$ using (\ref{2}).
Then replacing (\ref{1}) using (\ref{3}) and taking expectation with respect to $\tau_n$ and $\tau_s$, we obtain:
\begin{align}
	&\hspace{4.5mm}\mathbb{E}[R_T \mid\hat a^* = a^*]\nonumber\\
    &\leq \mu^*T - {\mu^*G(b,t) + \mu_{min}+L_{a^*}\Delta_{max}\over 1+G(b,t)}\mathbb{E}[\tau_s] + {\big(G(b,t)+L_{a^*}\big)\Delta_{max}\over 1+G(b,t)}\mathbb{E}[\tau_n] - n\sum_{i\in A}{\mu_i-\mu_{min}\over \mu_i} \nonumber\\&\hspace{4.5mm}- \mathbb{E}[\Gamma_T - \Gamma_{\tau_s}\mid\hat a^* = a^*]\nonumber\\
    &= \mu^*\mathbb{E}[\tau_s] - {\mu^*G(b,t) + \mu_{min}+L_{a^*}\Delta_{max}\over 1+G(b,t)}\mathbb{E}[\tau_s] + {\big(G(b,t)+L_{a^*}\big)\Delta_{max}\over 1+G(b,t)}\mathbb{E}[\tau_n] - n\sum_{i\in A}{\mu_i-\mu_{min}\over \mu_i} \nonumber\\&\hspace{4.5mm}+ \mu^*\big(T-\mathbb{E}[\tau_s]\big) - \mathbb{E}[\Gamma_T - \Gamma_{\tau_s}\mid\hat a^* = a^*]\nonumber\\
    &= {\Delta_{max}(1-L_{a^*})\over 1+G(b,t)}\mathbb{E}[\tau_s-\tau_n] + \Delta_{max}\cdot\mathbb{E}[\tau_n] - n\sum_{i\in A}{\mu_i-\mu_{min}\over \mu_i} + \mathbb{E}[R_T - R_{\tau_s}\mid\hat a^* = a^*].\label{2v2}
\end{align}
Then, the evaluation of $\mathbb{E}[R_T \mid\hat a^* = a^*]$ boils down to evaluating $\mathbb{E}[\tau_n]$, $\mathbb{E}[\tau_s-\tau_n]$ and $\mathbb{E}[R_T - R_{\tau_s}\mid \hat a^* = a^*]$.
We obtain from Lemma~\ref{dominance} and (\ref{2v2}) that
\begin{align*}
	&\hspace{4.5mm}\mathbb{E}[R_T \mid\hat a^* = a^*]\\
	&\leq \cfrac{\Delta_{max}(1-L_{a^*})}{1+G(b,t)}\cdot\cfrac{\big(n\cdot \sum_{i\in A}{1+G(b,t)\over G(b,t)\mu_i}-\cfrac{2n}{\mu^*}\big)\big(G(b,t)+1\big)}{G(b,t)-1} + \Delta_{max}n\sum_{i\in A}\cfrac{1+G(b,t)}{G(b,t)\mu_i} - n\sum_{i\in A}\cfrac{\mu_i-\mu_{min}}{\mu_i} \\&\hspace{4.5mm}+ \mathbb{E}[R_T - R_{\tau_s}\mid\hat a^* = a^*]\\
	&= n\bigg[\cfrac{\big(G(b,t)-L_{a^*}\big)\big(G(b,t)+1\big)}{G(b,t)\big(G(b,t)-1\big)}\sum_{a\in A}\cfrac{\Delta_{max}}{\mu_a}-\cfrac{a\Delta_{max}(1-L_{a^*})}{\mu^*\big(G(b,t)-1\big)}-\sum_{a\in A}\cfrac{\mu_a-\mu_{min}}{\mu_a}\bigg] + \mathbb{E}[R_T - R_{\tau_2}\mid\hat a^* = a^*] \\
&\overset{\text{(i)}}{\leq} n\bigg[\cfrac{2\big(G(b,t)-L_{a^*}\big)}{G(b,t)-1}\sum_{a\in A}\cfrac{\Delta_{max}}{\mu_a}\bigg] + \mathbb{E}[R_T - R_{\tau_2}\mid\hat a^* = a^*] \\
	&= O(\log T) + \mathbb{E}[R_T - R_{\tau_2}\mid\hat a^* = a^*],
\end{align*}
where (i) follows because $G(b,t)+1<2G(b,t)$.
By leveraging Eqs (\ref{t1}) and (\ref{t2}), the expected accumulative payment $\mathbb{E}[B_T]$ can also be upper bounded by
\begin{equation*}
	\mathbb{E}[B_T] = b\cdot(\mathbb{E}[\tau_n]+\mathbb{E}[\tau_s-\tau_n]) \leq \sum_{a\neq a^*}{2b(G(b,t)+1)\over \mu_a(G(b,t)-1)}\cdot q\ln T = O(\log T).
\end{equation*}

Next, for simplicity, we consider a system with $A=\lbrace 1, 2\rbrace$, where $\mu_1 > \mu_2$ and $\theta_1, \theta_2>0$.
The idea of the policy is that the agent keeps offering payment $b$ to the users pulling arm $1$ to help accumulate reward from arm $1$ and keep the arm in the leading side, i.e., arm $1$ generates at least half of accumulative reward, until time step $\tau_s$ when arm $1$ dominates and has an overwhelming chance to be the only arm that can generate rewards after monopoly happens.
This phenomenon is formulated as follows: suppose at time step $\tau_s$, $S_1(\tau_s)+S_2(\tau_s)=n_0$, and $S_2(\tau_s)=u_0n_0$ with $0<u_0<{1\over 2}$ and $u_0n_0\gg\theta_1, \theta_2$.
We estimate the probability of a ``bad'' event $D(u_0, n_0)$, where at some time step $t'>\tau_s$ we have $S_1(t')+S_2(t')=n>n_0$ and $S_2(t')\geq un$ with $0<u_0<u<{1\over 2}$, by leveraging the improved exponential embedding method, $D(u_0, n_0)$ can be expressed as follows:
\begin{equation*}
    D(u_0, n_0) =  \bigg(\sum_{i=u_0n_0}^{un-1}r_2(i)<\sum_{i=n_0-u_0n_0}^{n-un-1}r_1(i) \bigg).
\end{equation*}
We will show later that $\mathbb{P}(D(u_0, n_0))$ is very small, and with $u_0n_0$ getting larger, $\mathbb{P}(D(u_0, n_0))$ is getting exponentially smaller.
This result is formally stated as follows:
\begin{lemma}
\label{dominate-rate}
Suppose at time step $\tau_s$ there are $n_0$ accumulative reward with $u_0n_0, 0<u_0<{1\over 2}$ generated by arm $2$.
Then, there exists a constant $\gamma\in(0,1/4)$, such that for any $u_0 < u < {1\over 2}$ and all large enough $n_0$, it holds that:
\begin{equation*}
    \mathbb{P}\bigg(\exists n > n_0, D(u_0, n_0)\bigg) \leq e^{-(u_0n_0)^\gamma}.
\end{equation*}
\end{lemma}

By the above lemma, with $u_0n_0 = O(\tau_n) = O(\log T)$, we get $\mathbb{P}\big(D(u_0, n_0)\big) = O(e^{-{(\log T)}^\gamma})$.
This result can be extended to the case with arm number $m\geq 2$, by viewing the sum of accumulative reward generated from all sub-optimal arms as the accumulative reward generated from a single ``super arm.''

Next, we bound the last part $\mathbb{E}[R_T - R_{\tau_s}\mid \hat a^* = a^*]$.
Note that the regret comes from pullings of sub-optimal arms, and the expected number of attempts for each arm to get a unit reward is $O(1)$ since $\mu_i>0, i\in A$.
Let $n_0$ denote the accumulative reward from all arms at time step $\tau_s$ with $u_0n_0, 0<u_0<{1\over 2}$ rewards generated by sub-optimal arms.
Note that $u_0n_0=O(\log T)$ since $u_0n_0<\tau_s$ and $\tau_s=O(\log T)$.
Then, by Lemma \ref{dominate-rate}, for the unit reward generated right after $\tau_s$, it is generated by sub-optimal arms with probability smaller than or equal to $e^{-(u_0n_0)^\gamma}$ with $\gamma\in(0, {1\over 4})$. When a unit reward is generated by sub-optimal arms, the probability that the next unit reward is also generated by sub-optimal arms is smaller than or equal to $e^{-(u_0n_0+1)^\gamma}$.
Thus, we can upper bound the expected regret $\mathbb{E}[R_T - R_{\tau_s}\mid \hat a^* = a^*]$ by
\begin{align}
	\mathbb{E}[R_T - R_{\tau_s}\mid \hat a^* = a^*]
	&\leq e^{-(u_0n_0)^\gamma} + e^{-(u_0n_0+1)^\gamma} + \cdots\nonumber\\
	&\leq \int_{u_0n_0-1}^\infty e^{-n^\gamma}dn\nonumber\\
	&= Ce^{-(u_0n_0-1)^\gamma},\label{fv1}
\end{align}


where $C$ only depends on $u_0n_0$ and $\gamma$ such that $C=O\big((u_0n_0)^{1-\gamma}\big)$ with $\gamma\in(0, 1/4)$.
Thus Eq.~(\ref{fv1}) is $o(\log T)$.
Now we get the expected regret up to time step $T$ as $\mathbb{E}[R_T] = O(\log T)$, this completes the proof.

\subsection{Proof of Lemma \ref{best-arm-rate-greedy}}
\begin{fact}[Chernoff-Hoeffding bound]
Let $Z_1, \cdots, Z_n$ be independent bounded random variables with $Z_i\in [a, b]$ for all $i$, where $-\infty < a \leq b < \infty$. Then for all $s\geq0$
\begin{equation*}
    \mathbb{P}\bigg(\bigg| {1\over n}\sum_{i=1}^n (Z_i-\mathbb{E}[Z_i]) \bigg|\geq s \bigg) \leq \exp\bigg(-{2ns^2\over (b-a)^2} \bigg).
\end{equation*}
\end{fact}

\noindent Let sequences $\lbrace X_i(t)\rbrace$ denote the Bernoulli reward with support $\lbrace 0, 1\rbrace$ generated by arm $i\neq a^*$ at time step $t$.
Thus, for each time step $t$, $X_i(t)$ is an i.i.d. random variable and $\mathbb{E}[X_i(t)] = \mu_i$.
At time step $\tau_n$, by the policy, each arm has at least $n$ accumulative reward.
Since $S_i(\tau_n)$ is the accumulative reward generated by arm $i$ at time step $\tau_n$ we have $S_i(\tau_n) \geq n$.
By Chernoff-Hoeffding bound, at time step $\tau_n$ for arm $i$, we get the following:
\begin{equation*}
    \mathbb{P}\bigg(\hat\mu_i(\tau_n) > \mu_i + {\Delta_i\over 2}\bigg)\leq 2e^{-2\mathbb{E}[T_i(\tau_n)]({\Delta_i\over 2})^2}= 2e^{-2{\mathbb{E}[S_i(\tau_n)]\over\mu_i}({\Delta_i\over 2})^2}\leq 2e^{-{n\Delta_i^2\over 2\mu_i}}.
\end{equation*}
The proof for arm $a^*$ also follows from similar arguments and thus is omitted for brevity.

\subsection{Proof of Lemma \ref{dominate-rate}}
Suppose at some time step $t$, there are $n$ accumulative reward from both arms.
Recall that for arm $i\in A$, $\sum_{j=n}^\infty r_i(j) < \infty$ and $\mathbb{E}\big[\sum_{j=n}^\infty r_i(j)\big] = \sum_{j=n}^\infty {1\over \mu_iF(j+\theta_i)}$ converges.
To prove Lemma \ref{dominate-rate}, we use the following lemma
\begin{lemma}
\label{ineq}
There exists a constant $n_0$ such that for all $n>n_0$,
\begin{equation*}
    \mathbb{P}\bigg( \bigg| {\sum_{j=n}^\infty r_i(j)\over \mathbb{E}\big[\sum_{j=n}^\infty r_i(j)\big]} - 1 \bigg| > n^{-{1\over 4}} \bigg) \leq e^{-n^{1\over 4}}, i\in A.
\end{equation*}
\end{lemma}
\noindent Given a constant $t$, define an event $E_{n_0}$ where the following conditions hold simultaneously:
\begin{align}
	\bigg| {\sum_{j=u_0n_0}^\infty r_2(j)\over \mathbb{E}\big[\sum_{j=u_0n_0}^\infty r_2(j)\big]} - 1 \bigg| &\leq (u_0n_0)^{-{1\over 4}},\label{4}\\
    \forall n>n_0, \bigg| {\sum_{j=un}^\infty r_2(j)\over \mathbb{E}\big[\sum_{j=un}^\infty r_2(j)\big]} - 1 \bigg| &\leq (un)^{-{1\over 4}},\label{5}\\
    \bigg| {\sum_{j=(1-u_0)n_0}^\infty r_1(j)\over \mathbb{E}\big[\sum_{j=(1-u_0)n_0}^\infty r_1(j)\big]} - 1 \bigg| &\leq \big((1-u_0)n_0\big)^{-{1\over 4}},\label{6}\\
    \forall n>n_0, \bigg| {\sum_{j=(1-u)n}^\infty r_1(j)\over \mathbb{E}\big[\sum_{j=(1-u)n}^\infty r_1(j)\big]} - 1 \bigg| &\leq \big((1-u)n\big)^{-{1\over 4}}.\label{7}
\end{align}
By Lemma \ref{ineq}, we obtain the probability of event $E_{n_0}$ as follows
\begin{equation*}
    \mathbb{P}(E_{n_0})\geq 1- 2e^{-(u_0n_0)^{1\over 4}} - \sum_{n>n_0}2e^{-(u_0n)^{1\over 4}} \geq 1-e^{-(u_0n_0)^\gamma},
\end{equation*}
with $\gamma\in(0, {1\over 4})$ depending only on $F$ and $u_0$.
If we show that for all large enough $u_0n_0$, $E_{n_0}\cap D(u_0, n_0) = 0$, then the proof is finished since it implies
\begin{equation*}
    \mathbb{P}\bigg(\exists n > n_0, D(u_0, n_0)\bigg) \leq \mathbb{P}(E_{n_0}^c) \leq e^{-(u_0n_0)^\gamma}.
\end{equation*}

We consider the definition of event $D(u_0, n_0)$. By (\ref{4})--(\ref{7}), we obtain
\begin{align*}
    \sum_{i=u_0n_0}^{un-1}r_2(i)
    &= \sum_{i=u_0n_0}^\infty r_2(i) - \sum_{i=un}^\infty r_2(i)\\
    &\geq \big(1+o(1)\big)\sum_{i=u_0n_0}^\infty{1\over \mu_2F(i+\theta_2)} - \big(1+o(1)\big)\sum_{i=un}^\infty{1\over \mu_2F(i+\theta_2)},
\end{align*}
and similarly,
\begin{equation*}
    \sum_{i=n_0-u_0n_0}^{n-un-1}r_1(i)
    \leq \big(1+o(1)\big)\sum_{i=(1-u_0)n_0}^\infty{1\over \mu_1F(i+\theta_1)} - \big(1+o(1)\big)\sum_{i=(1-u)n}^\infty{1\over \mu_1F(i+\theta_1)}.
\end{equation*}
By contradiction, suppose that $E_{n_0}\cap D(u_0, n_0) \neq 0$. It then follows that
\begin{align*}
    &\big(1+o(1)\big)\sum_{i=u_0n_0}^\infty{1\over \mu_2F(i+\theta_2)} - \big(1+o(1)\big)\sum_{i=un}^\infty{1\over \mu_2F(i+\theta_2)}\\
    &< \big(1+o(1)\big)\sum_{i=(1-u_0)n_0}^\infty{1\over \mu_1F(i+\theta_1)} - \big(1+o(1)\big)\sum_{i=(1-u)n}^\infty{1\over \mu_1F(i+\theta_1)},
\end{align*}
which implies
\begin{equation}
    \sum_{i=u_0n_0}^{(1-u_0)n_0}{1\over \mu_1F(i+\theta_1)}
    < \big(1+o(1)\big)\sum_{i=un}^{(1-u)n}{1\over \mu_1F(i+\theta_1)}\label{8}.
\end{equation}
We want to show that (\ref{8}) cannot hold as $u_0n_0$ goes large, which implies $E_{n_0}\cap D(u_0, n_0) = 0$.
Since $F(x)=\Omega(x^\alpha)$, there exists $k>0$ such that
\begin{align*}
    \sum_{i=un}^{(1-u)n}{1\over \mu_1F(i+\theta_1)}
    &\leq k\bigg({n_0\over n}\bigg)^\alpha \sum_{i=un}^{(1-u)n}{1\over \mu_1F({n_0\over n}i+{n_o\over n}\theta_1)}\\
    &= k\bigg({n_0\over n}\bigg)^\alpha \sum_{i=un_0}^{(1-u)n_0}{1\over \mu_1F(i+\theta_1)}.
\end{align*}
Also, note that $[un_0, (1-u)n_0]\subset [u_0n_0, (1-u_0)n_0]$.
Therefore, there exists a constant $d\in(0, 1)$ such that
\begin{equation*}
    \sum_{i=un}^{(1-u)n}{1\over \mu_1F(i+\theta_1)}
    \leq dk\bigg({n_0\over n}\bigg)^\alpha\sum_{i=u_0n_0}^{(1-u_0)n_0}{1\over \mu_1F(i+\theta_1)},
\end{equation*}
which contradicts with (\ref{8}) since $o(1)$ goes to $0$ as $u_0n_0$ goes to infinity, and this completes the proof.

\subsection{Proof of Lemma \ref{ineq}}
Let $R_n = \sum_{j=n}^\infty r_i(j)$, $h(j) = \mu_iF(j+\theta_i)$, $Z_n = \sum_{j=n}^\infty {1\over h(j)^2}$.
We first show that for any $t\in\mathbb{R}^+$, we have
\begin{align}
    \mathbb{P}(R_n - \mathbb{E}[R_n]>t\sqrt{Z_n})&\leq e^{-t},\label{4v2}
\end{align}
and
\begin{align}
	\mathbb{P}(R_n - \mathbb{E}[R_n]<-t\sqrt{Z_n})&\leq e^{-t}.\label{5v2}
\end{align}

We only prove the first inequality and the proof of the second one is similar.
Given a constant $s$, we have:
\begin{align}
    \mathbb{P}(R_n - \mathbb{E}[R_n] > t\sqrt{Z_n})
    &\overset{\text{(i)}}{=} \mathbb{P}\bigg(e^{s(R_n - \mathbb{E}[R_n])} > e^{st\sqrt{Z_n}}\bigg)\nonumber\\
    &\overset{\text{(ii)}}{\leq} e^{-st\sqrt{Z_n}}\mathbb{E}\bigg[e^{s\sum_{j\geq n}(r_i(j)-{1\over h(j)})} \bigg]\nonumber\\
    &= e^{-st\sqrt{Z_n}}\prod_{j\geq n}\mathbb{E}\bigg[e^{s(r_i(j)-{1\over h(j)})} \bigg]\nonumber\\
    &\overset{\text{(iii)}}{=} e^{-st\sqrt{Z_n}}\prod_{j\geq n}{e^{-{s\over h(j)}}\over 1-{s\over h(j)}}\nonumber\\
    &= e^{-st\sqrt{Z_n}}\prod_{j\geq n}e^{{-s\over h(j)}}\bigg[1+{s\over h(j)}+{{s^2\over h(j)^2}\over 1-{s\over h(j)}} \bigg]\nonumber\\
    &\overset{\text{(iv)}}{\leq} e^{-st\sqrt{Z_n}}\prod_{j\geq n}e^{2s^2\over h(j)^2}\nonumber\\
    &\leq \exp(2s^2Z_n - st\sqrt{Z_n}),\label{9}
\end{align}
where (i) follows from multiplying both sides by a variable $s$ and exponentiate both sides, (ii) follows from Markov's inequality, (iii) is because given random variable $X\sim Exp(\lambda)$, $\mathbb{E}[e^{aX}]={1\over 1-{a\over \lambda}}, a<\lambda\nonumber$, and (iv) follows from $e^x \geq 1+x$.
We set $s = {1\over \sqrt{Z_n}}$, which is achievable since there exists $n$ such that ${1\over \sqrt{Z_n}} \leq {h(n)\over 2}$.
Thus, by (\ref{9}), we obtain $\mathbb{P}(R_n - \mathbb{E}[R_n]>t\sqrt{Z_n})\leq e^{-t}$.
Next, we use Lemma 1 in \citet{oliveira2009onset}, which is restated as follows:
\begin{lemma}[\citet{oliveira2009onset}, Lemma 1]
\label{Oliveira}
	Define a feedback function 
	$F(x) = \Theta(x^\alpha)$ where $\alpha>1$,
	and define the quantity
	\begin{equation*}
		S_r(n) = \sum_{j=n}^\infty{1\over F(j)^r}, r\in\mathbb{R}^+, n\in\mathbb{N}.
	\end{equation*}
	Then, for all $r\geq 1$, $S_r(n)$ converges and as $n\rightarrow +\infty$
	\begin{equation*}
		S_r(n)\rightarrow {n\over (r\alpha-1)F(n)^r}.
	\end{equation*}
\end{lemma}
By using Lemma \ref{Oliveira}, we obtain $\sqrt{S_2(n)} = n^{-{1\over2}}S_1(n)$ asymptotically.
Note that $S_1(n) = \mu_i\mathbb{E}[R_n]$ and $S_2(n) = \mu_i^2Z_n$.
Therefore, we obtain the relation between $\mathbb{E}[R_n]$ and $\sqrt{Z_n}$ as $\sqrt{Z_n} = n^{-{1\over2}}\mathbb{E}R_n$ asymptotically.
Then we replace $t$ by $n^{1\over 4}$ in both (\ref{4v2}) and (\ref{5v2}), and we get the inequality in Lemma \ref{ineq}.

\section{Proof of Theorem \ref{thm-ucb-list}}
\RestateUCB*
We start in a similar way as the proof of Theorem \ref{thm-etc}.
By the law of total expectation, the expected regret up to $T$ can be bounded as follows:
\begin{align*}
    \mathbb{E}[R_T] &= \mathbb{E}[R_T \mid\hat a^* = a^*]\mathbb{P}(\hat a^* = a^*) + \mathbb{E}[R_T\mid\hat a^*\neq a^*]\mathbb{P}(\hat a^*\neq a^*)\\
    &\leq \mathbb{E}[R_T \mid\hat a^* = a^*] + T\cdot\mathbb{P}(\hat a^*\neq a^*).
\end{align*}
We want to bound both $\mathbb{E}[R_T \mid\hat a^* = a^*]$ and $\mathbb{P}(\hat a^*\neq a^*)$ to get the regret bound.
We first consider $\mathbb{E}[R_T \mid\hat a^* = a^*]$.
After decomposing, we have:
\begin{align}
    \mathbb{E}[R_T \mid\hat a^* = a^*]
    &= \mathbb{E}[R_{\tau_2} \mid\hat a^* = a^*] + \mathbb{E}[R_T - R_{\tau_2} \mid\hat a^* = a^*]\nonumber\\
    &= \mathbb{E}[R_{\tau_1}] + \mathbb{E}[R_{\tau_2} - R_{\tau_1} \mid\hat a^* = a^*] + \mathbb{E}[R_T - R_{\tau_2} \mid\hat a^* = a^*].\label{6v2}
\end{align}
Note that after initialization, i.e., let $t_0$ be the time step when initialization is finished, each arm $a$ has $T_a(t_0)\geq 1$ since the number of attempts for each arm $a$ to get a unit reward is a geometric random variable with parameter larger than ${G(b,t)\mu_a\over 1+G(b,t)}$, which is independent of time.
During the exploration phase, since the regret is caused by pullings of sup-optimal arms, the expected regret after $t$ time steps can be written as
\begin{equation*}
    \sum_{a\neq a^*, a\in A} \Delta_a\mathbb{E}[T_a(t)].
\end{equation*}
Thus we can bound the expected regret during the exploration phase $\mathbb{E}[R_{\tau_1}]$ by bounding each $\mathbb{E}[T_a(\tau_1)]$ for $a\neq a^*$.
Let $U(t)$ denote the set of arms that can get payment at time $t$.
Consider the following two cases during the exploration phase:

\textbf{(a)} At time $t\leq\tau_1$, $a^*\in U(t)$ and there exists at least one suboptimal arm $a\in A, a\neq a^*$ such that $a\in U(t)$.
Recall that $c_a(t)= \sqrt{\ln T/2T_a(t)}$ is the confidence bound of arm $a$ at time step.
In this case, we have:
\begin{align}
	\mathbb{P}\big(\exists a\neq a^*: a\in U(t), a^*\in U(t)\big)
	&\overset{\text{(i)}}{\leq} \mathbb{P}\big(\hat\mu_a(t)+c_a(t)>\hat\mu^*(t)-c_{a^*}(t)\big)\cdot \mathbb{P}\big(\hat\mu^*(t)+c_{a^*}(t)>\hat\mu_a(t)-c_a(t) \big)\nonumber\\
	&\leq \mathbb{P}\big(\hat\mu_a(t)+c_a(t)>\hat\mu^*(t)-c_{a^*}(t)\big)\nonumber\\
	&\overset{\text{(ii)}}{\leq} \mathbb{P}\bigg(\hat\mu_a(t)+c_a(t)>\mu_a+\cfrac{\Delta_a}{2}\bigg) + \mathbb{P}\bigg(\hat\mu^*(t)-c_{a^*}(t)<\mu^*-\cfrac{\Delta_a}{2}\bigg)\label{revised-1},
\end{align}
where (i) is obtained since arm $a, a^*\in U(t)$ implies that the upper confidence bound of both arms is larger than the other arms's lower confidence bound, (ii) is because $\mu_a+\Delta_a/2=\mu^*-\Delta_a/2$, and the event $\lbrace\hat\mu_a(t)+c_a(t)> \hat\mu^*(t)-c_{a^*}(t)\rbrace$ implies either $\lbrace\hat\mu_a(t)+c_a(t)> \mu_a+\Delta_a/2\rbrace$ or $\lbrace\hat\mu^*(t)< \mu^*-\Delta_a/2\rbrace$.
We consider the first probability in Eq.~(\ref{revised-1}).
By Chernoff-Hoeffding bound we have
\begin{align}
	\mathbb{P}\bigg(\hat\mu_a(t)+c_a(t)>\mu_a+\cfrac{\Delta_a}{2}\bigg)
	&= \mathbb{P}\bigg(\hat\mu_a(t)-\mu_a>\cfrac{\Delta_a}{2}-c_a(t)\bigg)\nonumber\\
	&\leq e^{-2T_a(t)\big({\Delta_a\over 2}-c_a(t)\big)^2}\nonumber\\
	&= e^{-\big(\ln T + {\Delta_a^2\over 2}T_a(t)-\Delta_a\sqrt{2T_a(t)\ln T}\big)}.\label{revised-2}
\end{align}
Let ${\Delta_a^2\over 2}T_a(t)-\Delta_a\sqrt{2T_a(t)\ln T}=0$, we obtain $T_a(t)=8\ln T/\Delta_a^2$ and Eq.~(\ref{revised-2}) equals $1/T$. Note that as $T_a(t)$ increases, Eq.~(\ref{revised-2}) decreases monotonically.
Similar bound can be obtained of the second probability in Eq.~(\ref{revised-1}).
Thus, in this case, the expected regret contributed by a suboptimal arm $a\in A$ is bounded by
\begin{align}
	\Delta_a\mathbb{E}[T_a(t)]
	&\leq \cfrac{8\ln T}{\Delta_a} + \Delta_aT\cdot\mathbb{P}\big(t<\tau_1:a\in U(t),a^*\in U(t)\big)\nonumber\\
	&\leq \cfrac{8\ln T}{\Delta_a} + 2\Delta_a.\label{fv2}
\end{align}

\textbf{(b)} At time $t\leq\tau_1$, $a^*$ is eliminated by some suboptimal arm $a\in U(t), a\neq a^*$.
In this case, with similar technique as that in case \textbf{(a)} and Chernoff-Hoeffding bound, we have
\begin{align*}
	\mathbb{P}\big(\exists a\neq a^*: a\in U(t), a^*\notin U(t)\big)
	&\leq \mathbb{P}\big(\hat\mu_a(t)-c_a(t)>\hat\mu^*(t)+c_{a^*}(t)\big)\\
	&\leq \mathbb{P}\big(\hat\mu_{a^*}(t)+c_{a^*}(t)\leq \mu_{a^*}-{\Delta_a\over 2}\big) + \mathbb{P} \big(\hat\mu_a(t)-c_a(t)\geq \mu_a + {\Delta_a\over 2}\big)\\
	&\leq e^{-2T_{a^*}(t)\big({\Delta_a\over 2}+c_{a^*}(t)\big)^2} + e^{-2T_a(t)\big({\Delta_a\over 2}+c_a(t)\big)^2}\\
	&= e^{-{\Delta_a^2\over 2}T_{a^*}(t)-\ln T-\Delta_a\sqrt{2T_{a^*}(t)\ln T}} + e^{-{\Delta_a^2\over 2}T_a(t)-\ln T-\Delta_a\sqrt{2T_a(t)\ln T}}\\
	&\leq 2T^{-1}.
\end{align*}
Note that $\mathbb{P}(\hat a^*\neq a^*)=\mathbb{P}\big(\exists a\neq a^*: a\in U(t), a^*\notin U(t)\big)$.
Thus, in this case the expected regret contributed by a suboptimal arm $a\in A$ is upper bounded by
\begin{equation}
	\Delta_a\mathbb{E}[T_a(t)]
	\leq \Delta_a T\cdot \mathbb{P}\big(a\in U(t), a^*\notin U(t)\big)=2\Delta_a.\label{fv3}
\end{equation}
Summing Eq.~(\ref{fv2}) and Eq.~(\ref{fv3}) over all suboptimal arms, the expected regret during the exploration phase is bounded by:
\begin{equation*}
	\mathbb{E}[R_{\tau_1}]
	\leq \sum_{a\neq a^*}\cfrac{8\ln T}{\Delta_a}+4\Delta_a.
\end{equation*}
During the exploration phase at time step $t<\tau_1$, since the agent offers payment $b$ to the user for pulling arm $i$, the probability that the arm $i$ is pulled is ${\lambda_i(t)+G(b,t)\over 1+G(b,t)}>{G(b,t)\over 1+G(b,t)}$.
Thus, the number of attempts for arm $i$ to get pulled is a geometric random variable with parameter at least ${G(b,t)\over 1+G(b,t)}$. Since the above cases \textbf{(a)} and \textbf{(b)} imply the requirement of ${8\ln T\over \Delta_a^2}+4$ expected number of pullings from suboptimal arms, thus, the expected number of pullings for a suboptimal arm $a$ to guarantee at most ${8\ln T\over \Delta_a^2}+4$ number of pullings on every suboptimal arm is upper bounded by:
\begin{equation*}
	\mathbb{E}[T_a(\tau_1)]\leq \cfrac{G(b,t)+1}{G(b,t)}\bigg(\cfrac{8\ln T}{\Delta_a^2}+4\bigg).
\end{equation*}
 Thus, $\mathbb{E}[\tau_1]$ is upper bounded by:
\begin{equation}
	\mathbb{E}[\tau_1]
	=\sum_{a\in A} \mathbb{E}[T_a(\tau_1)]
	\overset{\text{(i)}}{\leq} \cfrac{G(b,t)+1}{G(b,t)}\bigg(\cfrac{8\ln T}{\Delta_{min}^2}+\sum_{a\neq a^*}\big(\cfrac{8\ln T}{\Delta_a^2}+4\big) \bigg),\label{revised-3}
\end{equation}

where (i) is due to the requirement of $T_{a^*}(\tau_1)$ to be at most ${8\ln T\over \Delta_{min}^2}$, since the exploration phase stops once the sampled strongest suboptimal arm is eliminated.
By the definition of dominance, arm $\hat a^*$ is expected to dominate at time $t\geq \tau_1$ if 
\begin{equation*}
	\mu_{\hat a^*}\mathbb{E}[T_{\hat a^*}(t)] \geq \sum_{a\neq \hat a^*}\mu_a\mathbb{E}[T_a(t)].
\end{equation*}

Similar as that in the proof of Lemma~\ref{dominance}, after tightening the condition by narrowing the left-hand-side and amplifying the right-hand-side, we obtain the sufficient condition of dominance as follows:
\begin{align}
	&\mu_{\hat a^*}\mathbb{E}[T_{\hat a^*}(t)] \geq \sum_{a\neq \hat a^*}\mu_a\mathbb{E}[T_a(t)]\nonumber\\
	&\Rightarrow \mu_{\hat a^*}T_{\hat a^*}(\tau_1)+\mu_{\hat a^*}\mathbb{E}[T_{\hat a^*}(t)-T_{\hat a^*}(\tau_1)] \geq \sum_{a\neq \hat a^*}\mu_a T_a(\tau_1)+\sum_{a\neq \hat a^*}\mu_a\mathbb{E}[T_a(t)-T_a(\tau_1)]\nonumber\\
	&\Rightarrow\mu_{\hat a^*}\mathbb{E}[T_{\hat a^*}(t)-T_{\hat a^*}(\tau_1)] \overset{\text{(i)}}{\geq} \sum_{a\neq\hat a^*}\big(\cfrac{8\mu_a}{\Delta_a^2}\ln T+4\mu_a\big)+\sum_{a\neq \hat a^*}\mu_a\mathbb{E}[T_a(t)-T_a(\tau_1)]\nonumber\\
	&\Rightarrow\cfrac{\mu_{\hat a^*}G(b,t)\mathbb{E}[t-\tau_1]}{G(b,t)+1} \overset{\text{(ii)}}{\geq} \sum_{a\neq\hat a^*}\big(\cfrac{8\mu_a}{\Delta_a^2}\ln T+4\mu_a\big)+\cfrac{\max_{a\neq \hat a^*}\mu_a\mathbb{E}[t-\tau_1]}{G(b,t)+1}\nonumber\\
	&\Rightarrow\mathbb{E}[t-\tau_1] \overset{\text{(iii)}}{\geq} \cfrac{G(b,t)+1}{\mu_{\hat a^*}G(b,t)-\max\limits_{a\neq\hat a^*}\mu_a}\sum_{a\neq \hat a^*}\big(\cfrac{8\mu_a}{\Delta_a^2}\ln T+4\mu_a\big),\label{24}
\end{align}
where (i) is obtained since $T_{\hat a^*}(\tau_1)>0$, (ii) is because by incentivizing arm $\hat a^*$, we have $\hat \lambda_{\hat a^*}(t)\geq{G(b,t)\over G(b,t)+1}$ and $\hat\lambda_a(t)\leq{1\over G(b,t)+1}$ for $a\neq \hat a^*$, and (iii) is the rearrangement.
Since time $\tau_2$ is defined as the earliest time to reach dominance, we can upper bound $\mathbb{E}[\tau_2-\tau_1]$ by
\begin{equation}
	\mathbb{E}[\tau_2-\tau_1] \leq \cfrac{G(b,t)+1}{\mu_{\hat a^*}G(b,t)-\max\limits_{a\neq\hat a^*}\mu_a}\sum_{a\neq \hat a^*}\big(\cfrac{8\mu_a}{\Delta_a^2}\ln T+4\mu_a\big).\label{tau2}
\end{equation}
Thus, we can bound the regret during the exploitation phase $\mathbb{E}[R_{\tau_2} - R_{\tau_1} \mid\hat a^* = a^*]$ in (\ref{6v2}) by
\begin{align*}
	\mathbb{E}[R_{\tau_2} - R_{\tau_1} \mid\hat a^* = a^*]
	&\overset{\text{(i)}}{\leq} \cfrac{\Delta_{max}}{G(b,t)+1}\cdot\mathbb{E}[\tau_2-\tau_1]\\
	&\leq \sum_{a\neq a^*}\bigg(\cfrac{8\Delta_{max}}{\Delta_a^2(G(b,t)-1)}\log T + \cfrac{4\Delta_{max}}{G(b,t)-1}\bigg),
\end{align*}
where (i) follows because during the exploitation phase there is always a positive probability $\hat\lambda_a(t)$ which is at most ${1\over G(b,t)+1}$ to pull suboptimal arm $a$.
By using Eqs (\ref{revised-3}) and (\ref{tau2}), the expected accumulative payment $\mathbb{E}[B_T]$ can also be upper bounded by
\begin{align*}
	\mathbb{E}[B_T]
	&= (\mathbb{E}[\tau_1]+\mathbb{E}[\tau_s-\tau_1])\cdot b\\
	&\leq \cfrac{G(b,t)+1}{G(b,t)}\bigg(\cfrac{8b\ln T}{\Delta_{min}^2}+\sum_{a\neq a^*}\big(\cfrac{8b\ln T}{\Delta_a^2}+4b\big) \bigg) + \cfrac{G(b,t)+1}{\mu_{\hat a^*}G(b,t)-\max\limits_{a\neq\hat a^*}\mu_a}\sum_{a\neq \hat a^*}\bigg(\cfrac{8b\mu_a}{\Delta_a^2}\ln T+4b\mu_a\bigg)\\
	&\overset{\text{(i)}}{\leq} \cfrac{G(b,t)+1}{G(b,t)}\bigg(\cfrac{8b\ln T}{\Delta_{min}^2}+\sum_{a\neq a^*}\big(\cfrac{8b\ln T}{\Delta_a^2}+4b\big) \bigg) + \cfrac{G(b,t)+1}{G(b,t)-1}\sum_{a\neq \hat a^*}\bigg(\cfrac{8b}{\Delta_a^2}\ln T+4b\bigg)\\
	&= \cfrac{G(b,t)+1}{G(b,t)}\cdot \cfrac{8b\ln T}{\Delta_{min}^2} + \bigg(\cfrac{G(b,t)+1}{G(b,t)}+\cfrac{G(b,t)+1}{G(b,t)-1}\bigg)\cdot\sum_{a\neq \hat a^*}\bigg(\cfrac{8b}{\Delta_a^2}\ln T+4b\bigg)\\
	&\overset{\text{(ii)}}{\leq} \cfrac{2G(b,t)+1}{G(b,t)-1}\bigg[\cfrac{8b\ln T}{\Delta_{min}^2} + \sum_{a\neq a^*}\bigg(\cfrac{8b\log T}{\Delta_a^2}+4b\bigg)\bigg],
\end{align*}
where (i) follows from $\mu^*>\mu_a$ for $a\neq a^*$, and (ii) follows from rearranging of the coefficients containing $G(b,t)$.
The choice of $\tau_2$ is sufficient to make the sampled best arm dominate at time step $\tau_2$ and have overwhelming probability to stay in leading side in monopoly after $\tau_2$.
The proof is the same as that in the proof of Theorem \ref{thm-etc}.
Thus, the expected regret of the last part $\mathbb{E}[R_T - R_{\tau_2}\mid \hat a^* = a^*] = O((\log T)^{1-\gamma}e^{-(\log T)^\gamma})=o(\log T)$ with $\gamma\in(0, {1\over 4})$ and the proof is the same as that in the proof of Theorem \ref{thm-etc}.

The above results show that we get the expected regret up to time step $T$ as $\mathbb{E}[R_T] = O(\log T)$ with expected accumulative payment $\mathbb{E}[B_T]=O(\log T)$, which completes the proof.

\end{appendices}
